\documentclass[11pt]{article}

\usepackage{acl}
\usepackage{fix-cm}
\usepackage{times}
\usepackage{latexsym}
\usepackage{amsmath,amssymb}
\usepackage{booktabs}
\usepackage{array}
\usepackage{tabularx}
\usepackage{enumitem}

\usepackage[T1]{fontenc}
\usepackage[utf8]{inputenc}
\usepackage{CJKutf8}

\providecommand{\skellist}[1]{%
\begin{minipage}[t]{\linewidth}
\begin{itemize}[leftmargin=1.0em,itemsep=0.02em,topsep=0pt,parsep=0pt,partopsep=0pt]
#1
\end{itemize}
\end{minipage}
}
\usepackage{microtype}

\usepackage{graphicx}
\usepackage{pgfplots}
\pgfplotsset{compat=1.18}

\usepackage{soul}
\usepackage{amsmath,amssymb}
\usepackage{booktabs}
\usepackage{multirow}
\usepackage{array}
\usepackage{tabularx}
\usepackage{threeparttable}
\usepackage{xcolor}
\usepackage{colortbl}
\usepackage{enumitem}
\usepackage{url}
\usepackage{stfloats}
\usepackage{float}
\usepackage{longtable}
\usepackage{makecell}
\newcolumntype{L}[1]{>{\raggedright\arraybackslash}p{#1}}

\newcommand{\dataset}{\textsc{ChLogic}}
\newcommand{\yes}{\textsc{Yes}}
\newcommand{\no}{\textsc{No}}
\newcommand{\unk}{\textsc{Unknown}}

\title{\dataset{}: Evaluating Robustness of Logical Reasoning in Chinese Expressions}

\author{
Peixian Zhou$^1$, Yuxu Chen$^1$, Chaorui Zhang$^2$, Wei Han$^2$, Bo Bai$^2$, Xueyan Niu$^{2}$\thanks{Corresponding author: niuxueyan3@huawei.com} \\
\smallskip
$^1$ College of Mathematics, Sichuan University, China \\
\smallskip
$^2$ Theory Lab, 2012 Labs, Huawei Technologies Co., Ltd.
}

\begin{document}
\maketitle

\begin{abstract}
Large language models perform increasingly well on standardized logical reasoning benchmarks, but whether this ability remains robust beyond English is unclear. We introduce \dataset{}, an English--Chinese aligned benchmark that tests whether models preserve logical reasoning performance when the same latent logical structure is expressed in English and diverse Chinese surface realizations.
Built from formal logical templates, the benchmark contains three data sets: (i) the General aligned set, derived from 60 General Propositions across nine template families; (ii) the Difficult aligned set, derived from 40 Difficult Problems; and (iii) the Chinese-only set, covering 15 language-specific phenomenon types. Each aligned item pairs one English reference expression with five Chinese realizations.
Experiments on Qwen3, Ministral, and GLM models reveal a persistent English--Chinese performance gap. Back-translation from standard Chinese into English often improves performance on the General aligned set, but produces mixed effects on the Difficult aligned set, where Qwen3-32B and GLM-5.1 perform worse after translation. These results indicate that Chinese surface realization, translation artifacts, and model-specific behavior jointly affect multilingual logical reasoning. Overall, \dataset{} provides a useful stress test for the robustness of multilingual reasoning.

\end{abstract}

\begin{figure*}[t]
\centering
\includegraphics[width=0.9\textwidth]{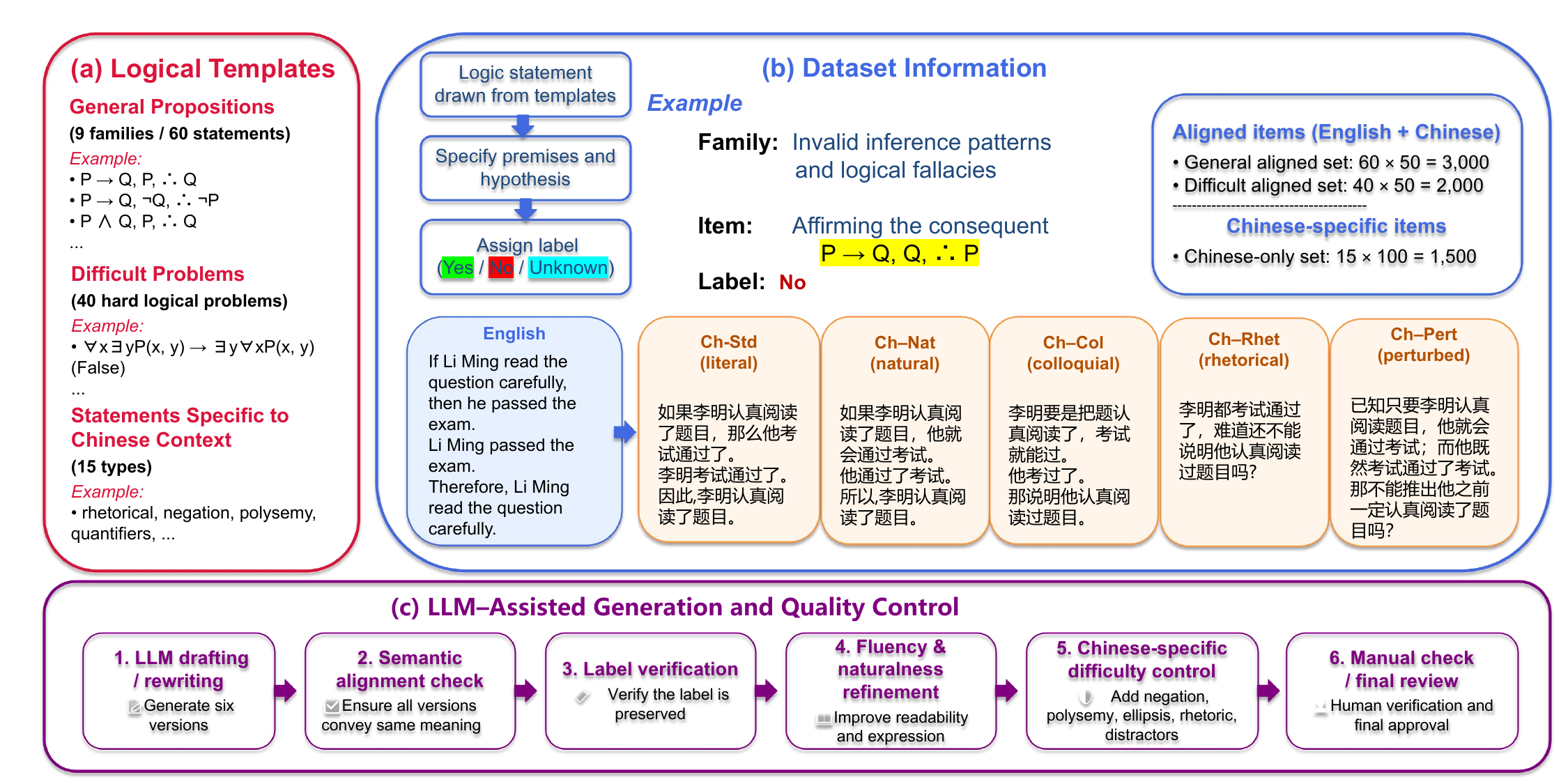}
\caption{\textbf{Workflow of} \dataset{} \textbf{benchmark construction.} The three stages are logical-template design, dataset composition, and LLM-assisted generation with quality control. Hanyu Pinyin transliterations and English renderings of the displayed Chinese example are provided in Appendix~\ref{app:figure_romanization}.}
\label{fig:pipeline}
\end{figure*}

\section{Introduction}
Logical reasoning is a central dimension of natural language understanding and has motivated datasets for natural language inference, rule following, symbolic reasoning, proof generation, and reading comprehension with logical constraints \citep{bowman2015snli,williams2018mnli,clark2020ruletaker,tafjord2021proofwriter,han2022folio,liu2020logiqa,yu2020reclor}. These benchmarks have improved model diagnostics, but their linguistic surface is predominantly English, template-like, or explicitly signaled by connectives such as \emph{if}, \emph{only if}, \emph{unless}, \emph{all}, \emph{some}, and \emph{no}. High performance under such conditions does not imply that a model can recover the same logical structure from Chinese expressions that are elliptical, rhetorical, idiomatic, multi-sense, colloquial, or pragmatically indirect.

Chinese creates a useful stress test for separating \emph{reasoning over a recovered structure} from \emph{recovering the structure itself}. The same logical relation takes many syntactic and pragmatic forms; negation and double negation are flexible; necessary and sufficient conditions are easily confused by markers such as \begin{CJK*}{UTF8}{gbsn}只要\end{CJK*} (as long as), \begin{CJK*}{UTF8}{gbsn}只有\end{CJK*} (only if), \begin{CJK*}{UTF8}{gbsn}除非\end{CJK*} (unless), and \begin{CJK*}{UTF8}{gbsn}否则\end{CJK*} (otherwise); and quantifier scope varies with word order, focus, and vague quantifiers. Rhetorical questions, irony, idioms, brand names, homophones, and omitted subjects obscure whether a conclusion is entailed, contradicted, or merely possible.

We construct \dataset, an English--Chinese aligned benchmark built from formal logical templates.
To narrow the gap between English-centric benchmarks and real-world Chinese reasoning, we use this English--Chinese logical alignment as a controlled framework. Our method extends existing monolingual robustness approaches, which test models on canonical and perturbed versions of the same input, to test whether models produce consistent logical judgments for equivalent problems expressed in different languages and surface realizations (Figure~\ref{fig:pipeline}). This design provides interpretable diagnostics, since it holds the logical label fixed while varying the linguistic realization. This is useful because multilingual logical reasoning data are rarely constructed with explicit alignment between surface-realization difficulty and underlying structure.

\begin{figure*}[t]
\centering
\includegraphics[width=\textwidth]{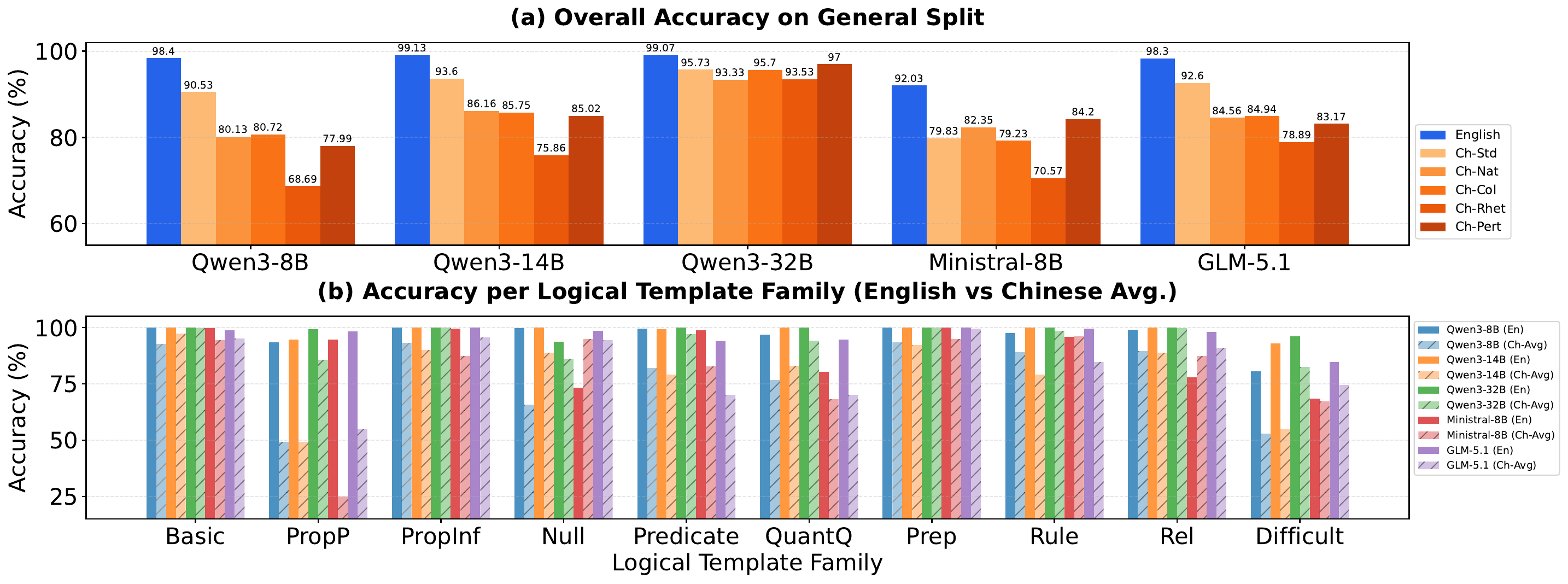}
\caption{\textbf{Main accuracy results on} \dataset{}. \textbf{(a)} Accuracy on the General aligned set for the English expression and five Chinese surface realizations: standard (Ch-Std), natural written (Ch-Nat), colloquial (Ch-Col), rhetorical-question (Ch-Rhet), and perturbed (Ch-Pert). \textbf{(b)} English accuracy with Chinese-average accuracy per logical-template family and for the Difficult aligned set; Chinese-average denotes the mean across the five Chinese variants.
}
\label{fig:main_results}
\end{figure*}

We evaluated Qwen3, Ministral, and GLM models on \dataset. As shown in Figure~\ref{fig:main_results}, the results show a persistent gap between logical reasoning abilities in English and Chinese. For example, GLM-5.1 achieves 98.30\% accuracy on English questions in the General aligned set, but drops to 78.89\% on rhetorical Chinese variants; in the Difficult aligned set, its accuracy falls from 84.70\% in English to 52.30\% on rhetorical Chinese. We construct back-translation probes by translating Chinese expressions back into English. Back-translation often improves performance on the General aligned set, but its effect is mixed on the Difficult aligned set and the Chinese-only set. These results are consistent with Chinese surface realization contributing to some errors, while also showing that translation can simplify, alter, or obscure information relevant to the judgment. Therefore, \dataset{} can be used as a critical stress test to evaluate the robustness of logical reasoning in multilingual models.

\section{Benchmark Construction}
\label{sec:benchmark_construction}

Figure~\ref{fig:pipeline} depicts the workflow of \dataset{} construction. Each instance is generated from a logical template and expressed through six semantically aligned surface realizations: one English reference form and five Chinese variants (standard, written, colloquial, rhetorical-question, and perturbed Chinese). All variants preserve the same logical structure while systematically increasing linguistic and pragmatic complexity, allowing us to evaluate whether models can maintain consistent reasoning across different forms of Chinese expression.

As shown in Table~\ref{tab:dataset_summary}, \dataset{} comprises three subsets. The \textit{General aligned set} is generated from 60 General Propositions across nine logical-template families, and the \textit{Difficult aligned set} is generated from 40 Difficult Problems. Both contain English--Chinese aligned items. The \textit{Chinese-only set} contains 15 language-specific phenomenon types, with 100 examples per type. Construction counts, quality-control responsibilities, and model-assisted stages are documented in Appendices~\ref{app:construction_audit} and~\ref{app:instance_expansion}.

\begin{table}[t]
\centering
\small
\setlength{\tabcolsep}{2.5pt}
\renewcommand{\arraystretch}{1.08}

\begin{tabular*}{\columnwidth}{
    @{\extracolsep{\fill}}
    l
    l
    r
    l
    @{}
}
\toprule
Data set & Inventory & Size & Forms \\
\midrule
General aligned
    & 60 templates
    & 3,000
    & En + 5 Ch \\
Difficult aligned
    & 40 skeletons
    & 2,000
    & En + 5 Ch \\
Chinese-only
    & 15 types
    & 1,500
    & Ch only \\
\bottomrule
\end{tabular*}

\caption{Composition of \dataset{}.
Details are provided in
Appendices~\ref{app:construction_details},
\ref{app:difficult_skeletons}, and
\ref{app:cn_examples}.}
\label{tab:dataset_summary}
\end{table}

The two aligned sets use a binary entailment--validity judgment with labels \yes{} and \no{}. The label \yes{} indicates that the stated conclusion is validly supported by the premises, whereas \no{} indicates either that the conclusion is contradicted or that the proposed inference is logically invalid, even when the conclusion itself is not explicitly contradicted. The Chinese-only set uses the same two labels plus \unk{} for genuine underdetermination, where the available information supports neither the conclusion nor its negation and no invalid inference is explicitly proposed. Appendix~\ref{app:label_inventory} reports the label inventory and distribution for each subset.

For example, in an affirming-the-consequent pattern (\(P \rightarrow Q,\; Q \nRightarrow P\)), the correct aligned-set label is \no{} because the proposed inference is invalid. By contrast, \unk{} is used in the Chinese-only set when the linguistic context leaves the target genuinely unresolved rather than presenting an identifiable invalid inference.

\subsection{Logical Templates}
\label{sec:templates_main}

\paragraph{General Propositions}
General Propositions contains 60 propositions organized into nine logical-template families, designed to provide a compact yet broad coverage of logical reasoning phenomena commonly studied in prior logic-oriented benchmarks. It consolidates logical forms that have been widely used in datasets such as LogicAsker, LogicBench, RuleTaker, ProofWriter, FOLIO, and LogicNLI, while introducing additional structures that are particularly relevant to Chinese linguistic variation and surface-to-logical-form normalization. 

The details of General Propositions are shown in Appendix~\ref{app:construction_details}:
(1) The first four families cover propositional structure: \textit{basic connectives}, \textit{propositional equivalence laws}, \textit{valid inference rules}, and \textit{invalid inference patterns}. These families test whether models preserve truth-functional relations and reject plausible but invalid conclusions such as affirming the consequent, denying the antecedent, illicit conversion, and affirming a disjunct;
(2) The next three families move to predicate logic: basic \textit{unary and binary predicates}, \textit{quantifier equivalence laws}, and \textit{predicate-level inference rules}. These examples test universal and existential scope, quantifier negation, invalid quantifier distribution, universal instantiation, existential generalization, and named-entity reasoning;
(3) The final two families cover \textit{multi-step rule chains} and \textit{relational logic}, including transitivity, non-transitivity, symmetry, and antisymmetry.

\paragraph{Difficult Problems}
Difficult Problems contains 40 mathematical-logic problems. These problems cover quantifier alternation, invalid quantifier exchange, relational properties, uniqueness, equality, and countermodel construction. When realized as natural-language examples, they are converted into the same judgment format as the rest of the benchmark. The appendix records their original formula-level labels under standard first-order semantics with a non-empty domain, while the main benchmark asks models to judge whether the stated conclusion follows from the given premises. The full list of the 40 difficult logical skeletons is provided in Appendix~\ref{app:difficult_skeletons}, Table~\ref{tab:app_difficult_40}.

\paragraph{Statements Specific to Chinese Context}
The Chinese-only set targets Chinese phenomena that are not well captured by direct translation of English logical templates. We construct 15 Chinese-only groups, each with 100 examples, covering condition markers, partial versus universal negation, ``not necessarily'' versus ``necessarily not'', multiple negation, rhetorical questions, omitted arguments, polysemy, segmentation ambiguity, vague quantifiers, comparison, temporal--causal confusion, concession and contrast, irony, idioms, rule-style Chinese, and homophone or brand-name shifts.
These groups are included because Chinese often encodes logical relations through compact markers, word order, discourse particles, or pragmatic force. For example, \begin{CJK*}{UTF8}{gbsn}只要\end{CJK*} A \begin{CJK*}{UTF8}{gbsn}就\end{CJK*} B (if A, then B) usually expresses that $A$ is sufficient for $B$, whereas \begin{CJK*}{UTF8}{gbsn}只有\end{CJK*} A \begin{CJK*}{UTF8}{gbsn}才\end{CJK*} B (B only if A) expresses that $A$ is necessary for $B$. Similarly, \begin{CJK*}{UTF8}{gbsn}不都\end{CJK*} (not all) and \begin{CJK*}{UTF8}{gbsn}都不\end{CJK*} (none; all not) differ in negation scope, and rhetorical questions may invert the literal polarity. Such cases test surface-to-logical-form normalization rather than formal inference alone. Representative examples are listed in Appendix~\ref{app:cn_examples}.

\subsection{LLM-Assisted Generation and Quality Control}
The benchmark is constructed using a template-first workflow. We first specify the logical templates, gold labels, premises, and target questions. LLMs generate surface realizations, producing one English reference form and multiple Chinese variants from the same specified logical content. The logical structure and gold labels are determined by the templates and never assigned by the models.

Quality control focuses on semantic alignment and label preservation. After generation, each realization is verified against its logical template to ensure the premises, target question, and gold label remain unchanged. We remove or revise an item if a realization introduces premises, omits information needed for judgment, changes the target question, alters the intended label, or creates ambiguity. For Chinese-specific phenomenon types, we verify that the target phenomenon is correctly instantiated and that the answer does not depend on world knowledge beyond the text.

LLMs assist surface generation, verification, revision, and back-translation, but do not assign labels or make final acceptance decisions. Rule-based checks identify malformed or inconsistent outputs, while human reviewers determine whether items are accepted, revised, regenerated, or discarded. Details are provided in Appendices~\ref{app:construction_audit} and~\ref{app:prompts}.

\section{Experimental Setup}
\label{sec:experimental_setup}
We evaluate binary entailment--validity judgment (\textsc{Yes}/\textsc{No}) on the General and Difficult aligned sets, and ternary judgment (\textsc{Yes}/\textsc{No}/\textsc{Unknown}) on the Chinese-only set. We evaluate the Qwen3 series, Ministral models, and GLM-5.1. All models are tested in a zero-shot setting with the same prompt and deterministic decoding. Model outputs are normalized into the valid label inventory for the relevant subset, and outputs that cannot be mapped to a valid label are counted as incorrect.

We report accuracy by data set, surface-realization condition, logical-template family, and phenomenon type. We also conduct Chinese-to-English back-translation probes: the standard Chinese field is translated back into English and evaluated with the same protocol. These probes test whether errors are due mainly to formal inference or to recovering the intended logical form from Chinese surface expressions. The complete prompt and parsing protocol are in Appendix~\ref{app:prompts}.

\section{Results}
\label{sec:results}

\begin{table*}[t]
\centering
\small
\setlength{\tabcolsep}{3.5pt}
\begin{tabular}{llrrrrrr}
\toprule
Model & Aligned set & English & Ch-Std & Ch-Nat & Ch-Col & Ch-Rhet & Ch-Pert \\
\midrule
Qwen3-0.6B & General & 78.3 & 78.3 & 78.3 & 78.3 & 78.3 & 78.3 \\
Qwen3-0.6B & Difficult & 55 & 55 & 55 & 55 & 55 & 55 \\
\midrule
Qwen3-8B & General & 98.4 & 90.53 & 80.13 & 80.72 & 68.69 & 77.99 \\
Qwen3-8B & Difficult & 80.5 & 52.5 & 61.58 & 59.48 & 34 & 56.28 \\
Qwen3-14B & General & 99.13 & 93.6 & 86.16 & 85.75 & 75.86 & 85.02 \\
Qwen3-14B & Difficult & 93.02 & 58.28 & 60.42 & 63.12 & 40.22 & 51.62 \\
Qwen3-32B & General & 99.07 & 95.73 & 93.33 & 95.7 & 93.53 & 97 \\
Qwen3-32B & Difficult & 96.05 & 83.1 & 87.8 & 86.15 & 69.35 & 85.2 \\
\midrule
Ministral-3B & General & 59.27 & 41.68 & 40.88 & 30.02 & 31.82 & 37.17 \\
Ministral-3B & Difficult & 73.3 & 47.9 & 49.02 & 49.06 & 46.6 & 46.3 \\
Ministral-8B & General & 92.03 & 79.83 & 82.35 & 79.23 & 70.57 & 84.2 \\
Ministral-8B & Difficult & 68.55 & 64.55 & 77.05 & 72.4 & 49.25 & 72.55 \\
\midrule
GLM-5.1 & General & 98.3 & 92.6 & 84.56 & 84.94 & 78.89 & 83.17 \\
GLM-5.1 & Difficult & 84.7 & 81 & 82.5 & 78.7 & 52.3 & 78.05 \\
\bottomrule
\end{tabular}
\caption{Overall accuracy (\%) on the binary General and Difficult aligned sets. Ch-Std, Ch-Nat, Ch-Col, Ch-Rhet, and Ch-Pert denote standard Chinese, natural written Chinese, colloquial Chinese, rhetorical-question Chinese, and perturbed Chinese.}
\label{tab:overall}
\end{table*}

\begin{table}[t]
\centering
\scriptsize
\setlength{\tabcolsep}{2.8pt}
\begin{tabular}{lrrrrrr}
\toprule
Model & G-En & G-Ch & G-BT & D-En & D-Ch & D-BT \\
\midrule
Qwen3-8B & 98.40 & 90.53 & 99.10 & 80.50 & 52.50 & 74.60 \\
Qwen3-14B & 99.13 & 93.60 & 98.90 & 93.02 & 58.28 & 81.75 \\
Qwen3-32B & 99.07 & 95.73 & 99.30 & 96.05 & 83.10 & 79.00 \\
Ministral-8B & 92.03 & 79.83 & 91.20 & 68.55 & 64.55 & 67.65 \\
GLM-5.1 & 98.30 & 92.60 & 97.73 & 84.70 & 81.00 & 62.95 \\
\bottomrule
\end{tabular}
\caption{Back-translation summary. G and D denote the General and Difficult aligned sets; Ch is standard Chinese; BT is the English back-translation of standard Chinese. Full family-level results are in Appendix~\ref{app:additional_results}.}
\label{tab:bt_summary}
\end{table}

Table~\ref{tab:overall} and Figure~\ref{fig:main_results} show an English--Chinese gap. Strong models perform near-perfectly on general English, with Qwen3-32B reaching 99.07\% and GLM-5.1 reaching 98.30\%. Accuracy drops under Chinese surface variation. GLM-5.1 falls to 78.89\% on general rhetorical Chinese and 83.17\% on general perturbed Chinese. The Difficult aligned set amplifies this pattern: Qwen3-32B scores 96.05\% on difficult English, but only 69.35\% on difficult rhetorical Chinese. Since aligned variants share the same structure and gold labels, these drops are consistent with difficulty in recovering the intended logical form from Chinese surface realizations, although they do not isolate one causal source.
\paragraph{Scaling and model profiles.} Scaling improves Chinese robustness but does not remove the problem. Within Qwen3, average accuracy across five Chinese variants rises from 79.61\% for Qwen3-8B to 95.06\% for Qwen3-32B, with largest gains on rhetorical and perturbed Chinese. The Difficult set improves with scale, but Qwen3-32B still drops from 96.05\% on difficult English to 69.35\% on difficult rhetorical Chinese. This suggests that larger models reduce but do not eliminate Chinese surface-normalization errors. Ministral-8B shows a different profile: although its general English score trails the stronger Qwen3 models and GLM-5.1, it remains competitive on several difficult Chinese variants. Thus, English-standard logical accuracy and Chinese-expression robustness are related but not identical abilities.
\paragraph{Template-level difficulty and fallacies.} Appendix~\ref{app:detailed} reports template-level results. Errors concentrate in forms that require global scope tracking, truth-condition comparison, or resistance to pragmatically tempting but invalid conclusions. Propositional equivalence is a bottleneck: Ministral-8B obtains 94.67\% on English equivalence, but only 23.56\% on standard Chinese and 8.00\% on rhetorical Chinese. Invalid inference is also diagnostic. Qwen3-8B drops from 99.75\% on English invalid inference to 41.50\% on rhetorical Chinese invalid inference, indicating difficulty with surface realizations that encourage fallacies such as affirming the consequent, denying the antecedent, illicit conversion, and existential fallacy. Appendix Table~\ref{tab:english_aug_fallacy} further shows that making compact English fallacy patterns more explicit can improve accuracy, suggesting that some fallacy errors are surface-realization failures rather than purely abstract reasoning failures.



\begin{table}[t]
\centering
\scriptsize
\setlength{\tabcolsep}{2.5pt}
\renewcommand{\arraystretch}{1.15}

\begin{tabularx}{\columnwidth}{
    @{}
    >{\raggedright\arraybackslash}p{0.22\columnwidth}
    >{\raggedright\arraybackslash}p{0.32\columnwidth}
    >{\raggedright\arraybackslash}X
    @{}
}
\toprule
\textbf{Error source}
&
\textbf{Typical trigger}
&
\textbf{Observed effect}
\\
\midrule

\makecell[tl]{Condition\\direction}
&
\makecell[tl]{
\begin{CJK*}{UTF8}{gbsn}只要/只有/除非\end{CJK*}
}
&
Confuses sufficient and necessary conditions
\\
\addlinespace[2pt]

Negation scope
&
\makecell[tl]{
\begin{CJK*}{UTF8}{gbsn}并非/不一定/未必\end{CJK*}
}
&
Reverses or weakens the conclusion
\\
\addlinespace[2pt]

\makecell[tl]{Rhetorical\\polarity}
&
\makecell[tl]{
\begin{CJK*}{UTF8}{gbsn}难道/怎么会\end{CJK*}
}
&
Treats rhetorical questions literally
\\
\addlinespace[2pt]

Quantifier scope
&
\makecell[tl]{
\begin{CJK*}{UTF8}{gbsn}有些/并非所有\end{CJK*}
}
&
Collapses universal and existential readings
\\
\addlinespace[2pt]

\makecell[tl]{Invalid\\inference}
&
\makecell[tl]{
converse/inverse\\[-1pt]
traps
}
&
Accepts plausible but invalid conclusions
\\
\addlinespace[2pt]

Ellipsis
&
\makecell[tl]{
omitted subjects\\[-1pt]
or objects
}
&
Adds unstated premises
\\

\bottomrule
\end{tabularx}

\caption{Representative Chinese surface phenomena behind common
logical-judgment failures. Detailed logical-template-family-level
results are provided in Appendix~\ref{app:detailed}.}
\label{tab:error_sources_main}
\end{table}

\paragraph{Back-translation diagnostics.} Back-translation probes further separate reasoning failures from Chinese surface-normalization failures. On the General aligned set, translating standard Chinese back into English often brings performance closer to the English-reference level: Qwen3-8B improves from 90.53\% to 99.10\%, Qwen3-32B from 95.73\% to 99.30\%, and GLM-5.1 from 92.60\% to 97.73\% (Appendix Table~\ref{tab:std_backtranslation}). The Difficult aligned set is more mixed: Qwen3-8B and Qwen3-14B improve substantially, while Qwen3-32B and GLM-5.1 decrease after back-translation (Appendix Table~\ref{tab:difficult_backtranslation}). 
For Chinese-specific statements, back-translation often helps on ellipsis, temporal-order versus causality, fuzzy quantifiers, and not-all versus all-not distinctions, but can hurt irony, homophone-based shifts, and some comparative structures (Appendix Tables~\ref{tab:phenomena_chinese}--\ref{tab:phenomena_backtranslation}). 
Overall, models often reason better once the structure is made explicit, but Chinese logical understanding still depends on a fragile normalization step.


\section{Analysis}
\label{sec:analysis}
The results support three conclusions: (1) Strong English logical performance does not guarantee robust Chinese understanding; (2) Scaling helps, but a sizable English--Chinese gap persists; (3) Model families differ in strengths and weaknesses across languages and surface realizations.

One interpretation is that \textsc{ChLogic} requires two reasoning stages. The model must first normalize a surface expression into a stable logical form, then perform inference over it. Existing logic benchmarks often emphasize the second stage.
In contrast, \textsc{ChLogic} stresses the first by holding the logical template fixed while varying the Chinese realization. The English--Chinese gaps suggest many errors occur before formal inference: models may misread conditional particles, lose negation scope, interpret rhetorical questions literally, overuse commonsense plausibility, or accept invalid converses.
Back-translation provides diagnostic evidence for this interpretation: on the General aligned set, back-translated English accuracy is much higher than standard Chinese accuracy and close to English-reference accuracy for all five models. Appendix~\ref{app:backtranslation_audit} discusses why this transformation is not neutral and reports the semantic-consistency audit protocol.

The results also show fine-grained evaluation is necessary. Overall accuracy can hide opposite failure modes. A model may have high English accuracy but limited Chinese robustness, as in Qwen3-14B on the Difficult aligned set. Conversely, a model may obtain misleadingly high scores when answer bias matches the label distribution, as in Qwen3-0.6B.
Template-family-level and phenomenon-level results expose these cases and identify Chinese forms requiring improvement.


\section{Related Work}
\paragraph{Logical reasoning benchmarks.}
Synthetic and semi-synthetic reasoning tasks such as bAbI \citep{weston2016babi}, RuleTaker \citep{clark2020ruletaker}, ProofWriter \citep{tafjord2021proofwriter}, and FOLIO \citep{han2022folio} control the logical structure and support proof-style evaluation; categorical-syllogism studies further examine template coverage and quantifier-related failures \citep{zong2024categorical}. Reading-comprehension benchmarks such as LogiQA and ReClor focus on logical reasoning in natural language passages \citep{liu2020logiqa,yu2020reclor}. These resources are valuable, but they do not directly test whether a model can maintain the same logical judgment when the surface realization is transformed into complex Chinese.

\paragraph{Chinese and multilingual evaluation.}
Chinese NLU benchmarks such as CLUE \citep{hu2020clue} and broad Chinese evaluation suites such as C-Eval \citep{huang2023ceval} measure many aspects of language understanding, knowledge, and reasoning. Cross-lingual and multilingual benchmarks include XNLI, MGSM, SeaEval, and MultiNRC \citep{conneau2018xnli,shi2023multilingual,wang2023seaeval,fabbri2025multinrc}. However, these resources do not isolate the same latent logical template across controlled English and Chinese realizations.

\paragraph{Robustness under linguistic variation.}
Contrast sets, behavioral testing, symbolic-template variation, and typographical perturbations show that high average accuracy can mask brittle reasoning \citep{gardner2020contrastsets,ribeiro2020checklist,mirzadeh2024gsmsymbolic,gan2024typos}. \dataset{} specializes this principle for logic: Each expression family is aligned by a formal template, and the diagnostic question is whether the judgment remains correct under Chinese realization changes.

\section{Conclusion}
We introduced \dataset, a controlled English--Chinese benchmark for logical understanding under complex Chinese expressions. Experiments on Qwen3, Ministral, and GLM models show that high English logical accuracy does not guarantee robust Chinese understanding. Scaling improves many general Chinese variants, but rhetorical, perturbed, quantifier-sensitive, and equivalence-law expressions remain difficult. Back-translation often improves performance on the General aligned set, but is unstable on the Difficult aligned set and mixed for Chinese-specific phenomena. The results therefore support a cautious interpretation: Chinese surface realization contributes to some failures, while translation artifacts and model-specific decision behavior also affect outcomes.

\section*{Limitations}
\dataset{} is a diagnostic benchmark rather than a complete measurement of Chinese reasoning ability. Its aligned items are template-driven and may not cover all forms of discourse-level reasoning. The model set is limited to seven contemporary systems, and the results may change as model families are updated. Although labels are specified before surface generation and items are manually checked, generated Chinese variants may still contain artifacts. Finally, back-translation is itself a model-mediated transformation and can remove Chinese cues or introduce English-side ambiguity.


\bibliography{ChLogic}

@inproceedings{bowman2015snli,
  title={A large annotated corpus for learning natural language inference},
  author={Bowman, Samuel R. and Angeli, Gabor and Potts, Christopher and Manning, Christopher D.},
  booktitle={Proceedings of EMNLP},
  year={2015}
}

@inproceedings{williams2018mnli,
  title={A broad-coverage challenge corpus for sentence understanding through inference},
  author={Williams, Adina and Nangia, Nikita and Bowman, Samuel R.},
  booktitle={Proceedings of NAACL-HLT},
  year={2018}
}

@inproceedings{hu2020clue,
  title={{CLUE}: A Chinese Language Understanding Evaluation Benchmark},
  author={Xu, Liang and Hu, Hai and Zhang, Xuanwei and Li, Lu and Cao, Chenjie and Li, Yudong and Xu, Yechen and Sun, Kai and Yu, Dian and Yu, Cong and others},
  booktitle={Proceedings of COLING},
  year={2020}
}

@article{huang2023ceval,
  title={{C-Eval}: A Multi-Level Multi-Discipline Chinese Evaluation Suite for Foundation Models},
  author={Huang, Yuzhen and Bai, Yuzhuo and Zhu, Zhihao and Zhang, Junlei and Zhang, Jinghan and Su, Tangjun and Liu, Junteng and Lv, Chuancheng and Zhang, Yikai and Lei, Yao and Fu, Yao and Sun, Maosong and He, Junxian},
  journal={arXiv preprint arXiv:2305.08322},
  year={2023}
}

@inproceedings{clark2020ruletaker,
  title={Transformers as Soft Reasoners over Language},
  author={Clark, Peter and Tafjord, Oyvind and Richardson, Kyle},
  booktitle={Proceedings of IJCAI},
  year={2020}
}

@inproceedings{tafjord2021proofwriter,
  title={{ProofWriter}: Generating Implications, Proofs, and Abductive Statements over Natural Language},
  author={Tafjord, Oyvind and Dalvi, Bhavana and Clark, Peter},
  booktitle={Findings of ACL-IJCNLP},
  year={2021}
}

@inproceedings{han2022folio,
  title={{FOLIO}: Natural Language Reasoning with First-Order Logic},
  author={Han, Simeng and Schoelkopf, Hailey and Zhao, Yilun and Qi, Zhenting and Riddell, Martin and Benson, Luke and Sun, Lucy and Zubova, Ekaterina and Qiao, Yejin and Burtell, Matthew and others},
  booktitle={Proceedings of EMNLP},
  year={2022}
}

@inproceedings{liu2020logiqa,
  title={{LogiQA}: A Challenge Dataset for Machine Reading Comprehension with Logical Reasoning},
  author={Liu, Jian and Cui, Leyang and Liu, Hanmeng and Huang, Dandan and Wang, Yile and Zhang, Yue},
  booktitle={Proceedings of IJCAI},
  year={2020}
}

@inproceedings{yu2020reclor,
  title={{ReClor}: A Reading Comprehension Dataset Requiring Logical Reasoning},
  author={Yu, Weihao and Jiang, Zihang and Dong, Yanfei and Feng, Jiashi},
  booktitle={Proceedings of ICLR},
  year={2020}
}

@article{weston2016babi,
  title={Towards AI-complete question answering: A set of prerequisite toy tasks},
  author={Weston, Jason and Bordes, Antoine and Chopra, Sumit and Rush, Alexander M. and van Merrienboer, Bart and Joulin, Armand and Mikolov, Tomas},
  journal={arXiv preprint arXiv:1502.05698},
  year={2016}
}

@inproceedings{gardner2020contrastsets,
  title={Evaluating Models' Local Decision Boundaries via Contrast Sets},
  author={Gardner, Matt and Artzi, Yoav and Basmova, Victoria and Berant, Jonathan and Bogin, Ben and Chen, Sihao and Dasigi, Pradeep and Dua, Dheeru and Elazar, Yanai and Gottumukkala, Ananth and others},
  booktitle={Findings of EMNLP},
  year={2020}
}

@inproceedings{ribeiro2020checklist,
  title={Beyond Accuracy: Behavioral Testing of NLP Models with {CheckList}},
  author={Ribeiro, Marco Tulio and Wu, Tongshuang and Guestrin, Carlos and Singh, Sameer},
  booktitle={Proceedings of ACL},
  year={2020}
}

@inproceedings{wan2024logicasker,
  title = {{LogicAsker}: Evaluating and Improving the Logical Reasoning Ability of Large Language Models},
  author = {Wan, Yuxuan and Wang, Wenxuan and Yang, Yiliu and Yuan, Youliang and Huang, Jen-tse and He, Pinjia and Jiao, Wenxiang and Lyu, Michael R.},
  booktitle = {Proceedings of the 2024 Conference on Empirical Methods in Natural Language Processing},
  pages = {2124--2155},
  year = {2024},
  publisher = {Association for Computational Linguistics}
}

@inproceedings{parmar2024logicbench,
  title = {{LogicBench}: Towards Systematic Evaluation of Logical Reasoning Ability of Large Language Models},
  author = {Parmar, Mihir and Patel, Nisarg and Varshney, Neeraj and Nakamura, Mutsumi and Luo, Man and Mashetty, Santosh and Mitra, Arindam and Baral, Chitta},
  booktitle = {Proceedings of the 62nd Annual Meeting of the Association for Computational Linguistics (Volume 1: Long Papers)},
  pages = {13679--13707},
  year = {2024},
  address = {Bangkok, Thailand},
  publisher = {Association for Computational Linguistics},
  doi = {10.18653/v1/2024.acl-long.739},
  url = {https://aclanthology.org/2024.acl-long.739/}
}

@article{ontanon2022logicinference,
  title = {{LogicInference}: A New Dataset for Teaching Logical Inference to seq2seq Models},
  author = {Ontan{\'o}n, Santiago and Ainslie, Joshua and Cvicek, Vaclav and Fisher, Zachary},
  journal = {arXiv preprint arXiv:2203.15099},
  year = {2022},
  url = {https://arxiv.org/abs/2203.15099}
}

@inproceedings{tian2021logicnli,
  title = {Diagnosing the First-Order Logical Reasoning Ability Through {LogicNLI}},
  author = {Tian, Jidong and Li, Yitian and Chen, Weixiang and Xiao, Li and He, Hao and Jin, Yaohui},
  booktitle = {Proceedings of the 2021 Conference on Empirical Methods in Natural Language Processing},
  pages = {3738--3747},
  year = {2021},
  address = {Online and Punta Cana, Dominican Republic},
  publisher = {Association for Computational Linguistics},
  doi = {10.18653/v1/2021.emnlp-main.303},
  url = {https://aclanthology.org/2021.emnlp-main.303/}
}

@article{rabern2026logicskills,
  title = {{LogicSkills}: A Structured Benchmark for Formal Reasoning in Large Language Models},
  author = {Rabern, Brian and Mondorf, Philipp and Plank, Barbara},
  journal = {arXiv preprint arXiv:2602.06533},
  year = {2026},
  url = {https://arxiv.org/abs/2602.06533}
}

@book{enderton2001mathematical,
  title     = {A Mathematical Introduction to Logic},
  author    = {Enderton, Herbert B.},
  edition   = {2},
  year      = {2001},
  publisher = {Academic Press}
}

@book{mendelson2015introduction,
  title     = {Introduction to Mathematical Logic},
  author    = {Mendelson, Elliott},
  edition   = {6},
  year      = {2015},
  publisher = {CRC Press}
}

@inproceedings{conneau2018xnli,
  title     = {XNLI: Evaluating Cross-lingual Sentence Representations},
  author    = {Conneau, Alexis and Rinott, Ruty and Lample, Guillaume and Williams, Adina and Bowman, Samuel R. and Schwenk, Holger and Stoyanov, Veselin},
  booktitle = {Proceedings of the 2018 Conference on Empirical Methods in Natural Language Processing},
  pages     = {2475--2485},
  year      = {2018},
  publisher = {Association for Computational Linguistics},
  doi       = {10.18653/v1/D18-1269}
}

@inproceedings{shi2023multilingual,
  title     = {Language Models are Multilingual Chain-of-Thought Reasoners},
  author    = {Shi, Freda and Suzgun, Mirac and Freitag, Markus and Wang, Xuezhi and Srivats, Srinivasan and Vosoughi, Soroush and Chung, Hyung Won and Tay, Yi and Ruder, Sebastian and Zhou, Denny and Das, Dipanjan and Wei, Jason},
  booktitle = {The Eleventh International Conference on Learning Representations},
  year      = {2023}
}

@article{mirzadeh2024gsmsymbolic,
  title   = {{GSM-Symbolic}: Understanding the Limitations of Mathematical Reasoning in Large Language Models},
  author  = {Mirzadeh, Iman and Alizadeh, Keivan and Shahrokhi, Hooman and Tuzel, Oncel and Bengio, Samy and Farajtabar, Mehrdad},
  journal = {arXiv preprint arXiv:2410.05229},
  year    = {2024}
}

@article{wang2023seaeval,
  title   = {{SeaEval} for Multilingual Foundation Models: From Cross-Lingual Alignment to Cultural Reasoning},
  author  = {Wang, Bin and Liu, Zhengyuan and Huang, Xin and Jiao, Fangkai and Ding, Yang and Aw, Ai Ti and Chen, Nancy F.},
  journal = {arXiv preprint arXiv:2309.04766},
  year    = {2023}
}

@article{fabbri2025multinrc,
  title   = {{MultiNRC}: A Challenging and Native Multilingual Reasoning Evaluation Benchmark for {LLM}s},
  author  = {Fabbri, Alexander R. and Mares, Diego and Flores, Jorge and Mankikar, Meher and Hernandez, Ernesto and Lee, Dean and Liu, Bing and Xing, Chen},
  journal = {arXiv preprint arXiv:2507.17476},
  year    = {2025}
}

@inproceedings{zong2024categorical,
  title     = {Categorical Syllogisms Revisited: A Review of the Logical Reasoning Abilities of {LLM}s for Analyzing Categorical Syllogism},
  author    = {Zong, Shi and Lin, Jimmy},
  booktitle = {Proceedings of the 1st Workshop on NLP for Science (NLP4Science)},
  year      = {2024},
  publisher = {Association for Computational Linguistics}
}

@article{gan2024typos,
  title   = {Reasoning Robustness of {LLM}s to Adversarial Typographical Errors},
  author  = {Gan, Esther and Zhao, Yiran and Cheng, Liying and Mao, Yancan and Goyal, Anirudh and Kawaguchi, Kenji and Kan, Min-Yen and Shieh, Michael},
  journal = {arXiv preprint arXiv:2411.05345},
  year    = {2024}
}
\clearpage
\appendix

\section{Additional Benchmark Construction Details}
\label{app:construction_details}
\subsection{Logical Templates and General Propositions}
\label{sec:skeleton-inventory}
\begin{table*}[t]
\centering
\scriptsize
\renewcommand{\arraystretch}{2.08}
\setlength{\tabcolsep}{6.8pt}

\newcommand{\tsym}[1]{{\fontsize{6.6}{8.0}\selectfont\ensuremath{#1}}}

\begin{tabularx}{\textwidth}{
>{\raggedright\arraybackslash}p{0.23\textwidth}
>{\centering\arraybackslash}p{0.055\textwidth}
>{\raggedright\arraybackslash}X}
\toprule
\textbf{Logical-template Family} & \textbf{\#} & \textbf{Representative Formal Templates} \\
\midrule

Basic propositional connectives
& 7
& \skellist{
    \item Atomic proposition: \tsym{P}.
    \item Negation: \tsym{\neg P}.
    \item Binary connectives: \tsym{P\land Q}, \tsym{P\lor Q}, \tsym{P\oplus Q}.
    \item Conditional forms: \tsym{P\rightarrow Q}, \tsym{P\leftrightarrow Q}.
} \\

Propositional equivalence laws
& 9
& \skellist{
    \item Idempotence: \tsym{P\land P\leftrightarrow P}, \tsym{P\lor P\leftrightarrow P}.
    \item Commutativity and associativity for \tsym{\land} and \tsym{\lor}.
    \item Distributivity between \tsym{\land} and \tsym{\lor}.
    \item De Morgan's laws: \tsym{\neg(P\land Q)\leftrightarrow \neg P\lor \neg Q}; \tsym{\neg(P\lor Q)\leftrightarrow \neg P\land \neg Q}.
    \item Double negation: \tsym{\neg\neg P\leftrightarrow P}.
    \item Material implication: \tsym{P\rightarrow Q\leftrightarrow \neg P\lor Q}.
    \item Contraposition: \tsym{P\rightarrow Q\leftrightarrow \neg Q\rightarrow \neg P}.
    \item Biconditional equivalence.
} \\

Propositional inference rules
& 11
& \skellist{
    \item Modus ponens: \tsym{P\rightarrow Q,\, P \therefore Q}.
    \item Modus tollens: \tsym{P\rightarrow Q,\, \neg Q \therefore \neg P}.
    \item Hypothetical syllogism: \tsym{P\rightarrow Q,\, Q\rightarrow R \therefore P\rightarrow R}.
    \item Disjunctive syllogism: \tsym{P\lor Q,\, \neg P \therefore Q}.
    \item Addition, simplification, and conjunction introduction.
    \item Resolution: \tsym{P\lor Q,\, \neg P\lor R \therefore Q\lor R}.
    \item Proof by cases; biconditional introduction and elimination.
} \\

Invalid inference patterns and logical fallacies
& 8
& \skellist{
    \item Affirming the consequent: \tsym{P\rightarrow Q,\, Q \nRightarrow P}.
    \item Denying the antecedent: \tsym{P\rightarrow Q,\, \neg P \nRightarrow \neg Q}.
    \item Affirming a disjunct: \tsym{P\lor Q,\, P \nRightarrow \neg Q}.
    \item Illicit conversion: \tsym{P\rightarrow Q \nRightarrow Q\rightarrow P}.
    \item Existential fallacy, illicit major, illicit minor, and undistributed middle.
} \\

Basic predicate logic
& 5
& \skellist{
    \item Unary predicate: \tsym{P(a)}.
    \item Binary relation: \tsym{R(a,b)}.
    \item Universal quantification: \tsym{\forall x P(x)}, \tsym{\forall x(P(x)\rightarrow Q(x))}.
    \item Existential quantification: \tsym{\exists x P(x)}, \tsym{\exists x(P(x)\land Q(x))}.
    \item Unique existence: \tsym{\exists!x P(x)}.
} \\

Quantifier equivalence laws
& 5
& \skellist{
    \item Quantifier negation: \tsym{\neg\forall x P(x)\leftrightarrow \exists x\neg P(x)}.
    \item Existential negation: \tsym{\neg\exists x P(x)\leftrightarrow \forall x\neg P(x)}.
    \item Universal distribution over conjunction: \tsym{\forall x(P(x)\land Q(x))\leftrightarrow \forall x P(x)\land \forall x Q(x)}.
    \item Existential distribution over disjunction.
    \item Invalid distribution: \tsym{\exists x P(x)\land \exists x Q(x)\nRightarrow \exists x(P(x)\land Q(x))}.
    \item Quantifier movement.
} \\

Predicate-logic inference rules
& 7
& \skellist{
    \item Universal instantiation: \tsym{\forall x P(x)\therefore P(c)}.
    \item Existential generalization: \tsym{P(c)\therefore \exists x P(x)}.
    \item Invalid universal generalization: \tsym{P(c)\nRightarrow \forall x P(x)}.
    \item Predicate-level MP: \tsym{\forall x(P(x)\rightarrow Q(x)),\, P(a)\therefore Q(a)}.
    \item Predicate-level MT: \tsym{\forall x(P(x)\rightarrow Q(x)),\, \neg Q(a)\therefore \neg P(a)}.
    \item Predicate-level transitive reasoning.
    \item Existential-instance reasoning.
} \\

Multi-step reasoning
& 4
& \skellist{
    \item Linear chain: \tsym{P\rightarrow Q,\, Q\rightarrow R,\, R\rightarrow S,\, P\therefore S}.
    \item Conjunctive condition chain: \tsym{(P\land Q)\rightarrow R,\, A\rightarrow P,\, B\rightarrow Q,\, A,\, B\therefore R}.
    \item Branching inference: \tsym{P\rightarrow R,\, Q\rightarrow S,\, P\lor Q\therefore R\lor S}.
    \item Multi-premise conjunction reasoning.
} \\

Relational logic
& 4
& \skellist{
    \item Transitivity: \tsym{R(a,b),\, R(b,c),\, \forall x\forall y\forall z((R(x,y)\land R(y,z))\rightarrow R(x,z))\therefore R(a,c)}.
    \item Non-transitivity: \tsym{R(a,b),\, R(b,c)\nRightarrow R(a,c)}.
    \item Symmetry: \tsym{R(a,b),\, \forall x\forall y(R(x,y)\rightarrow R(y,x))\therefore R(b,a)}.
    \item Antisymmetry: \tsym{R(a,b),\, R(b,a),\, \forall x\forall y((R(x,y)\land R(y,x))\rightarrow x=y)\therefore a=b}.
} \\

\bottomrule
\end{tabularx}
\caption{The nine logical-template families of General Propositions used in our benchmark. The table reports each logical-template family, the number of propositions, and representative formal templates. The inventory contains 60 basic logical statements; after filtering and de-duplication, the current General aligned set contains 3,000 items.}
\label{tab:skeletons}
\end{table*}

Table~\ref{tab:skeletons} lists the 60 general propositions used in \dataset{}. The inventory is not copied from any single prior benchmark. Instead, it consolidates logical forms that are commonly tested in prior logic-oriented evaluations and supplements them with additional structures that are useful for testing Chinese surface-to-logical-form normalization. Prior benchmarks such as LogicAsker, LogicBench, LogicInference, RuleTaker, ProofWriter, FOLIO, LogicNLI, and LogicSkills cover many of the standard ingredients used here, including propositional connectives, equivalence laws, valid inference rules, common fallacies, first-order quantification, predicate-level inference, and multi-step rule chaining \citep{wan2024logicasker,parmar2024logicbench,ontanon2022logicinference,clark2020ruletaker,tafjord2021proofwriter,han2022folio,tian2021logicnli,rabern2026logicskills}. LogicAsker, for example, defines 34 atomic logical rules spanning propositional logic, predicate logic, and common fallacies, which provides one representative reference point for the benchmark-attested part of our inventory \citep{wan2024logicasker}. Our goal is therefore to build a compact but broad 60-proposition inventory that covers common benchmark-tested logical forms while adding structures that are especially relevant to Chinese linguistic variation.

The first three families cover core propositional logic. Basic propositional connectives include atomic propositions, negation, conjunction, disjunction, implication, and biconditionality, all of which are standard building blocks in prior logic benchmarks \citep{wan2024logicasker,parmar2024logicbench,ontanon2022logicinference}. We additionally include exclusive disjunction because Chinese expressions such as \begin{CJK*}{UTF8}{gbsn}要么……要么……\end{CJK*}(either\ldots or\ldots) and ordinary \begin{CJK*}{UTF8}{gbsn}或者\end{CJK*} (or) create a useful inclusive--exclusive ambiguity. Propositional equivalence laws include idempotence, commutativity, associativity, distributivity, De Morgan's laws, double negation, material implication, contraposition, and biconditional equivalence, most of which are directly aligned with equivalence-law diagnostics in prior work \citep{wan2024logicasker,parmar2024logicbench}. Propositional inference rules include modus ponens, modus tollens, hypothetical syllogism, disjunctive syllogism, addition, simplification, conjunction introduction, resolution, proof by cases, and biconditional introduction/elimination; these forms are widely used in rule-based and proof-style reasoning benchmarks \citep{clark2020ruletaker,tafjord2021proofwriter,wan2024logicasker,parmar2024logicbench}.

The fourth family contains invalid inference patterns and logical fallacies. This family includes affirming the consequent, denying the antecedent, affirming a disjunct, illicit conversion, existential fallacy, illicit major, illicit minor, and undistributed middle. Several of these fallacies are explicitly included in LogicAsker's diagnostic inventory \citep{wan2024logicasker}. We retain them because Chinese markers such as \begin{CJK*}{UTF8}{gbsn}只要\end{CJK*} (as long as), \begin{CJK*}{UTF8}{gbsn}只有\end{CJK*} (only if), \begin{CJK*}{UTF8}{gbsn}除非\end{CJK*} (unless), and \begin{CJK*}{UTF8}{gbsn}否则\end{CJK*} (otherwise) can easily induce converse or inverse interpretations, making fallacy recognition a central part of Chinese logical robustness.

The next three families cover first-order logic. The basic predicate-logic family includes unary predicates, binary predicates, universal quantification, and existential quantification, which are standard in FOL-oriented benchmarks such as LogicNLI, FOLIO, LogicBench, and LogicSkills \citep{tian2021logicnli,han2022folio,parmar2024logicbench,rabern2026logicskills}. We add unique existence because Chinese expressions such as \begin{CJK*}{UTF8}{gbsn}唯一\end{CJK*} (unique), \begin{CJK*}{UTF8}{gbsn}只有一个 \end{CJK*} (only one), and \begin{CJK*}{UTF8}{gbsn}恰好一个\end{CJK*} (exactly one) are common and logically precise but are not usually isolated as a separate benchmark family. The quantifier-equivalence family covers quantifier negation, existential negation, quantifier distribution, invalid quantifier distribution, and quantifier movement. These items are motivated by prior FOL validity tasks, but the Chinese versions emphasize scope ambiguity caused by word order and conditional placement \citep{wan2024logicasker,han2022folio,rabern2026logicskills}. The predicate-inference family includes universal instantiation, existential generalization, invalid universal generalization, predicate-level modus ponens and modus tollens, predicate-level transitive reasoning, and existential-instance reasoning. These templates test whether a model can preserve the difference between a named individual, an existential witness, and a universally quantified object.

The final two families test composition and relation tracking. Multi-step reasoning includes linear chains, conjunction-conditioned chains, branching inference, and multi-premise reasoning, following the broader motivation of rule-chain and proof-oriented benchmarks \citep{clark2020ruletaker,tafjord2021proofwriter,wan2024logicasker,parmar2024logicbench}. Relational logic includes transitivity, non-transitivity, symmetry, and antisymmetry. Although relation reasoning is part of first-order logical reasoning in prior work \citep{tian2021logicnli,han2022folio,rabern2026logicskills}, we make these relational properties explicit because Chinese word order, comparative structures, and passive-like descriptions can obscure argument direction.

Overall, the 60-proposition inventory combines benchmark-attested logical forms with additions introduced for Chinese logical robustness. The additions include exclusive disjunction, unique existence, invalid quantifier distribution, quantifier movement, invalid universal generalization, existential-instance tracking, branching and multi-premise discourse, non-transitivity, and antisymmetry. This design lets us test whether models can recover the same underlying logical form across English, standard Chinese, natural Chinese, colloquial Chinese, rhetorical Chinese, and perturbed Chinese surface realizations.

\subsection{Statements Specific to Chinese Context}

Beyond translating English logical templates, we additionally design 15 groups of statements specific to Chinese context that are frequent in real Chinese usage and likely to disrupt surface-to-logical-form normalization. The motivation is that Chinese often encodes logical relations through compact surface markers, discourse particles, word order, pragmatic tone, and context-dependent expressions rather than through fully explicit logical connectives. As a result, expressions that are superficially similar may correspond to substantially different logical structures. For example, if A, then B typically expresses that A is a sufficient condition for B, whereas only if A, then B expresses that A is a necessary condition for B. Although the two constructions share similar lexical material, they map to different implication directions. Similar contrasts also arise in partial versus universal negation, vague quantification, ellipsis, rhetorical questions, and concession structures. A model that relies mainly on surface lexical overlap may therefore recover the wrong logical form even when the sentence appears linguistically simple.

Based on this observation, we construct statements specific to Chinese context to test whether models can distinguish subtle form--meaning mismatches that are especially salient in Chinese. These phenomena include condition-marker ambiguity, where near-identical conditional phrases may reverse or shift the direction of implication; partial versus universal negation, where expressions such as not all and  all not / none encode different scopes of negation; double negation, where the surface presence of multiple negative markers may either strengthen or cancel negation depending on context; rhetorical questions, where the intended assertion must be inferred from interrogative form and discourse tone; omitted arguments, where the model must recover implicit subjects, objects, or predicates from context; polysemy, where the same word may trigger different predicate meanings; segmentation ambiguity, where different word boundaries yield different interpretations; vague quantifiers, where expressions such as \begin{CJK*}{UTF8}{gbsn}不少\end{CJK*} (many; not few), \begin{CJK*}{UTF8}{gbsn}大多数\end{CJK*} (most), or \begin{CJK*}{UTF8}{gbsn}有些\end{CJK*} (some) introduce non-exact quantificational force; comparison, where relational direction and comparison standard must be tracked; temporal-causal confusion, where temporal succession may be mistaken for causal implication; concession and contrast, where markers such as \begin{CJK*}{UTF8}{gbsn}虽然\end{CJK*} (although), \begin{CJK*}{UTF8}{gbsn}但是\end{CJK*} (but), and \begin{CJK*}{UTF8}{gbsn}倒是\end{CJK*} (however; on the contrary) affect the relation between clauses; irony, where the literal polarity may differ from the intended meaning; idioms, where the logical content is not compositionally recoverable from individual words; rule-style Chinese, where normative or policy-like expressions encode conditional obligations; and brand- or homophone-based concept shifts, where surface similarity can mislead the model into treating distinct concepts as identical.

Table~\ref{tab:cn_phenomena} summarizes these 15 groups of statements and their target difficulties. Each group currently contains 100 generated samples. Representative examples are provided in Appendix~\ref{app:cn_examples}. Overall, these groups of statements specific to Chinese context target cases where the underlying logical structure is not difficult in itself, but becomes more difficult to recognize because Chinese surface realization may obscure, shift, or pragmatically enrich the logical relation.

\subsection{LLM-Assisted Generation and Quality Control}
\label{sec:quality-control}

The construction follows a template-first procedure. We first specify the logical template, premises, hypothesis or target question, and label. LLMs are used as controlled surface-realization assistants rather than as unconstrained label generators. Given the pre-specified logical information, the generation prompt asks the model to produce one English standard expression and the five Chinese surface realizations while preserving the same latent logical relation and the same label.

Quality control focuses on semantic alignment and label preservation. We remove or rewrite an item if any realization adds a new premise, deletes information needed for the judgment, changes the target question, changes the label, or introduces uncontrolled ambiguity. We also revise Chinese surface realizations for fluency, naturalness, and intended difficulty. For statements specific to Chinese context, we additionally check that the target phenomenon is actually present and that the answer does not depend on private world knowledge outside the given text. Candidate items are screened by rule-based checks, while LLM-based verification is used as an auxiliary diagnostic and to propose revisions. Neither mechanism changes the locked template label. Final acceptance, revision, regeneration, or rejection requires human inspection. Appendix~\ref{app:construction_audit} provides the responsibility matrix and construction statistics.

The full prompts used for surface realization, verification, and model evaluation are provided in Appendix~\ref{app:prompts}. This prompt disclosure is intended to make the generation and evaluation protocol more reproducible while making clear that labels are specified before surface generation rather than inferred from model outputs.

\begin{table*}[t]
\centering
\scriptsize
\begin{tabular}{p{0.06\linewidth} p{0.32\linewidth} p{0.52\linewidth}}
\toprule
No. & Phenomenon & Target difficulty \\
\midrule
1 & Condition markers: as long as, only if, unless, otherwise &
Distinguishing sufficient conditions, necessary conditions, unless-conditions, and otherwise-clauses. \\

2 & Partial negation vs. universal negation &
Separating existential counterexamples from universal negative claims. \\

3 & ``Not necessarily'' vs. ``necessarily not'' &
Avoiding confusion between uncertainty and logical negation. \\

4 & Double or multiple negation &
Recovering the correct polarity and scope under nested negation. \\

5 & Rhetorical questions &
Interpreting rhetorical force as implicit affirmation, denial, or conditional exclusion. \\

6 & Omitted arguments &
Resolving omitted subjects, objects, conditions, or conclusions. \\

7 & Polysemy and character-level ambiguity &
Distinguishing repeated or identical surface realizations with different meanings. \\

8 & Segmentation ambiguity &
Handling ambiguity caused by Chinese word segmentation, attachment, and scope boundaries. \\

9 & Vague quantifiers &
Avoiding over-strengthening vague quantifiers such as ``mostly'', ``basically'', or ``not necessarily'' into universal claims. \\

10 & Comparison structures &
Tracking comparative direction, implicit standards, and non-transitive surface comparisons. \\

11 & Temporal order vs. causality &
Distinguishing temporal succession or correlation from causal entailment. \\

12 & Concession, contrast, and embedded conditions &
Composing concessive clauses, contrastive markers, and nested conditional rules. \\

13 & Irony &
Recovering intended polarity when literal meaning and speaker intent diverge. \\

14 & Idioms, proverbs, and internet slang &
Mapping conventionalized expressions to their implied logical consequences. \\

15 & Rule-style Chinese, brand names, abbreviations, and homophones &
Interpreting formal rule language and avoiding concept shifts caused by names, abbreviations, or homophones. \\
\bottomrule
\end{tabular}
\caption{Statements specific to Chinese context used in \dataset. The table summarizes the linguistic phenomena and their target surface-to-logical-form normalization difficulties. Representative examples are provided in Appendix~\ref{app:cn_examples}.}
\label{tab:cn_phenomena}
\end{table*}

\section{Construction Audit and Responsibility Matrix}
\label{app:construction_audit}

All prompts were fixed within each construction stage. DeepSeek-V3 was used for surface generation, revision, and back-translation, while Yi-1.5-34B-Chat served as an independent verifier. The evaluated Qwen3, Ministral, and GLM models did not participate in benchmark construction. Logical templates, premises, target questions, and labels were fixed before surface generation and were not modified by either construction model. 
 
\begin{table*}[t]
\centering
\scriptsize
\setlength{\tabcolsep}{3.2pt}
\begin{tabular}{
p{0.15\textwidth}
p{0.19\textwidth}
p{0.13\textwidth}
p{0.16\textwidth}
p{0.29\textwidth}}
\toprule
Stage
& Tool or annotator
& Flagged or rejected
& Revised or regenerated
& Final responsibility \\
\midrule

Surface generation
& DeepSeek-V3
& --
& --
& Produces candidate surface realizations only; the logical templates and labels remain fixed \\

LLM verification
& Yi-1.5-34B-Chat
& 300 flagged
& --
& Checks premise preservation, target-question consistency, label preservation, and unintended ambiguity \\

Revision proposal
& DeepSeek-V3
& --
& 300
& Revises or regenerates flagged linguistic realizations without changing the underlying templates or labels \\

Back-translation
& DeepSeek-V3
& --
& --
& Produces diagnostic English translations without altering the original items or benchmark labels \\

Human review
& Human annotators
& 0
& 100 manually revised
& Makes the final accept, revise, regenerate, or reject decision \\
\bottomrule
\end{tabular}

\caption{Construction stages and responsibility boundaries. Yi-1.5-34B-Chat flagged 300 candidate realizations for revision, which were subsequently revised or regenerated using DeepSeek-V3. During final human review, 100 additional items were manually revised, and no items were rejected.}
\label{tab:construction_audit}
\end{table*}

Human reviewers conducted the final review of all retained items, checking Chinese fluency, semantic alignment with the underlying template, premise and question preservation, label consistency, and unintended ambiguity. Problematic items were manually revised or returned for regeneration, and final acceptance decisions were made by the human reviewers.

\section{Instance Expansion, Filtering, and Release Structure}
\label{app:instance_expansion}

Each of the 60 General Propositions is instantiated into 50 scenario-level items, yielding $60\times50=3{,}000$ latent aligned items. Each of the 40 Difficult Problems is likewise instantiated into 50 scenario-level items, yielding $40\times50=2{,}000$ latent aligned items. Scenario instantiation varies entities, predicates, events, and application settings while preserving the formal relation and locked label. Every latent aligned item is then rendered as one English reference and five Chinese surface realizations: standard, natural written, colloquial, rhetorical-question, and perturbed Chinese. The counts 3,000 and 2,000 refer to latent aligned items; the six realizations are parallel evaluation conditions and are not counted as six independent logical items.

The Chinese-only set contains 15 phenomenon types with 100 items each, for a total of 1,500 items. It is generated directly in Chinese because its target phenomena depend on Chinese lexical, syntactic, pragmatic, or orthographic properties. \dataset{} is released as a test-only diagnostic benchmark and does not define train, development, or test partitions. Exact-string duplicate checking on the supplied 1,500-item Chinese-only file found no duplicate full-text items. Prompts for instantiation and realization appear in Appendix~\ref{app:prompts}.

\section{Label Inventory and Distribution}
\label{app:label_inventory}

The aligned sets evaluate binary entailment--validity judgments. They distinguish validly supported conclusions (\yes{}) from contradicted conclusions or explicitly proposed invalid inferences (\no{}). They do not include \unk{} because their templates are constructed to present a determinate validity judgment. The Chinese-only set additionally uses \unk{} because several Chinese-specific phenomena intentionally create genuine underdetermination through omitted arguments, vague quantification, pragmatic ambiguity, or unresolved scope.

\begin{table*}[t]
\centering
\small
\setlength{\tabcolsep}{5pt}
\begin{tabular}{lrrrrrrr}
\toprule
Subset & \yes{} & \yes{} (\%) & \no{} & \no{} (\%) & \unk{} & \unk{} (\%) & Total \\
\midrule
General aligned set & 2,350 & 78.33 & 650 & 21.67 & 0 & 0.00 & 3,000 \\
Difficult aligned set & 1,100 & 55.00 & 900 & 45.00 & 0 & 0.00 & 2,000 \\
Chinese-only set & 399 & 26.60 & 457 & 30.47 & 644 & 42.93 & 1,500 \\
\bottomrule
\end{tabular}
\caption{Gold-label counts and proportions by subset. The aligned sets are binary; the Chinese-only set is ternary.}
\label{tab:label_distribution}
\end{table*}

These distributions make aggregate accuracy sensitive to answer bias. For example, an almost-always-\yes{} strategy can obtain 78.33\% on the General aligned set without reliable reasoning, while a strong \no{} bias benefits more on the Difficult aligned set, where \no{} accounts for 45.00\%. Accordingly, the degenerate-model analysis in Appendix~\ref{app:additional_results} interprets accuracy together with label frequencies and logical-template-family behavior.

\section{Bias-Aware Metrics for the General Aligned Set}
\label{app:bias_metrics}

This section reports bias-aware metrics for Qwen3-0.6B and Ministral-3B on the General aligned set. These two models are selected because they exhibit the clearest answer-distribution biases. Each of the 60 General templates contributes 50 items, yielding 2,350 \yes{} and 650 \no{} instances.

Qwen3-0.6B predicts \yes{} for every item under all six surface-realization conditions, so its metrics are identical across conditions. For Ministral-3B, we report the English-reference and rhetorical-Chinese conditions. For both models, we present accuracy, balanced accuracy, Macro-F1, per-label precision, recall, and F1, confusion-matrix counts, and predicted-label distributions.

\begin{table*}[t]
\centering
\scriptsize
\setlength{\tabcolsep}{5pt}
\begin{tabular}{llrrrr}
\toprule
Model & Condition & Accuracy & Balanced Acc. & Macro-F1 & Predicted \yes{} \\
\midrule
Qwen3-0.6B & All six conditions & 78.33 & 50.00 & 43.93 & 100.00 \\
Ministral-3B & English reference & 59.27 & 67.71 & 56.89 & 45.13 \\
Ministral-3B & Rhetorical Chinese & 31.83 & 56.49 & 30.92 & 10.17 \\
\bottomrule
\end{tabular}
\caption{Bias-aware performance (\%) on the General aligned set. Predicted \no{} is the complement of Predicted \yes{}.}
\label{tab:general_bias_metrics}
\end{table*}

\begin{table*}[t]
\centering
\scriptsize
\setlength{\tabcolsep}{4.5pt}
\begin{tabular}{llrrrrrr}
\toprule
Model & Condition &
\yes{} Prec. & \yes{} Rec. & \yes{} F1 &
\no{} Prec. & \no{} Rec. & \no{} F1 \\
\midrule
Qwen3-0.6B & All six conditions
& 78.33 & 100.00 & 87.85
& 0.00 & 0.00 & 0.00 \\
Ministral-3B & English reference
& 91.65 & 52.81 & 67.01
& 32.62 & 82.62 & 46.78 \\
Ministral-3B & Rhetorical Chinese
& 100.00 & 12.98 & 22.98
& 24.12 & 100.00 & 38.86 \\
\bottomrule
\end{tabular}
\caption{Per-label precision, recall, and F1 (\%) on the General aligned set.}
\label{tab:general_per_label_metrics}
\end{table*}

\begin{table*}[t]
\centering
\scriptsize
\setlength{\tabcolsep}{5pt}
\begin{tabular}{llrrrr}
\toprule
Model & Condition &
Gold \yes{} $\rightarrow$ Pred. \yes{} &
Gold \yes{} $\rightarrow$ Pred. \no{} &
Gold \no{} $\rightarrow$ Pred. \yes{} &
Gold \no{} $\rightarrow$ Pred. \no{} \\
\midrule
Qwen3-0.6B & All six conditions & 2,350 & 0 & 650 & 0 \\
Ministral-3B & English reference & 1,241 & 1,109 & 113 & 537 \\
Ministral-3B & Rhetorical Chinese & 305 & 2,045 & 0 & 650 \\
\bottomrule
\end{tabular}
\caption{Confusion-matrix counts for the General aligned set.}
\label{tab:general_confusion_counts}
\end{table*}

These metrics reveal failure modes that aggregate accuracy obscures. Qwen3-0.6B attains 78.33\% accuracy because the General aligned set is \yes{}-majority, yet its balanced accuracy is only 50.00\% and its \no{} recall is 0. Its apparent performance therefore reflects a universal-\yes{} strategy rather than reliable discrimination between valid and invalid conclusions.

Ministral-3B exhibits the opposite bias. In the English-reference condition, it predicts \no{} for 54.87\% of items; under rhetorical Chinese, this rises to 89.83\%. Although its \no{} recall reaches 100\% in rhetorical Chinese, its \yes{} recall falls to 12.98\%, producing a low Macro-F1 of 30.92\%. Thus, high accuracy on \no{}-heavy template families should not be interpreted as uniformly strong logical reasoning. Together, the two models illustrate why balanced accuracy, per-label metrics, and confusion counts are necessary complements to aggregate accuracy. Gold-label distributions by subset and General template family are reported in Table~\ref{tab:label_distribution}.

\section{Back-Translation Audit and Interpretation}
\label{app:backtranslation_audit}

Back-translation is treated as a diagnostic transformation rather than a neutral removal of Chinese surface information. It may simplify wording, alter rhetorical force, change negation scope or condition direction, remove homophonic or culturally grounded cues, and introduce English-side ambiguity. DeepSeek-V3 was used to produce the back-translated English condition, as summarized in Table~\ref{tab:construction_audit}.

To assess semantic consistency, sampled Chinese items were compared with their back-translated English versions along four dimensions: preservation of the premises and target question, preservation of the locked label, preservation of negation scope and condition direction, and preservation of rhetorical or pragmatic force. Particular attention was paid to pragmatically dependent phenomena, including rhetorical questions, irony, and homophone-based shifts. Because back-translation may simplify wording, weaken pragmatic cues, or introduce English-side artifacts, performance changes are interpreted as diagnostic rather than causal evidence about Chinese-to-logical-form normalization.

\section{Additional Results and Analysis}
\label{app:additional_results}

Table~\ref{tab:overall} and Figure~\ref{fig:main_results} give the complete overall results for all evaluated models. The clearest global pattern is that English-standard logical accuracy is much easier than robust Chinese logical understanding. Among the non-degenerate strong models, Qwen3-32B obtains the highest general English score, 99.07\%, while GLM-5.1 is close at 98.30\%. However, most models still drop under Chinese surface variation. Qwen3-32B is substantially more robust than the smaller Qwen3 models, but GLM-5.1 still falls to 78.89\% on general rhetorical Chinese and 83.17\% on general perturbed Chinese. These drops are not due to different underlying logical structures, because the examples are aligned. They are consistent with the view that a major difficulty lies in normalizing Chinese surface expressions into the intended logical structure, rather than in formal inference alone.

The Difficult aligned set, visualized in Figure~\ref{fig:main_results}B, amplifies this separation. Qwen3-32B reaches 96.05\% on difficult English, but its difficult Chinese scores are 83.10\% on standard Chinese, 87.80\% on natural Chinese, 86.15\% on colloquial Chinese, 69.35\% on rhetorical Chinese, and 85.20\% on perturbed Chinese. GLM-5.1 shows the same broad English--Chinese contrast, though at lower Difficult aligned set accuracy. The particularly low rhetorical scores indicate that rhetorical polarity, pragmatic implication, and implicit negation remain difficult even for the strongest models in the study.

Ministral-8B has a different profile. Its general English score is 92.03\%, below Qwen3-8B, Qwen3-14B, Qwen3-32B, and GLM-5.1, but it is competitive on several Chinese variants. On the General aligned set, it reaches 82.35\% on natural Chinese and 84.20\% on perturbed Chinese. On the Difficult aligned set, it is the best model on natural Chinese (77.05\%), colloquial Chinese (72.40\%), rhetorical Chinese (49.25\%), and perturbed Chinese (72.55\%). This indicates that English-standard logical skill and Chinese-expression robustness are separable: a model can be less accurate on canonical English but more robust to some Chinese realization styles.

Ministral-3B behaves atypically. Its general English score is only 59.27\%, and its five Chinese general scores range from 30.02\% to 41.68\%, indicating weak general logical judgment. Yet on the English condition of the Difficult aligned set it rises to 73.30\%, higher than its own general English result by 14.03 points. Its difficult Chinese scores, around 46--49\%, are also higher than or comparable to its general Chinese scores. This violates the expected ordering in which the Difficult aligned set should be harder than the General aligned set. As discussed below, this is best understood as a label-bias and answer-distribution artifact rather than a genuine advantage on harder logic.

\subsection{Scaling trend within Qwen3}

The Qwen3 results in Table~\ref{tab:overall} and Figure~\ref{fig:main_results} summarize the scaling trend. Ignoring Qwen3-0.6B as a degenerate baseline, scaling from 8B to 32B improves general Chinese robustness: the average of the five general Chinese variants rises from 79.61\% for Qwen3-8B to 85.28\% for Qwen3-14B and 95.06\% for Qwen3-32B. The largest gains occur in rhetorical and perturbed Chinese: rhetorical Chinese increases from 68.69\% to 93.53\%, and perturbed Chinese increases from 77.99\% to 97.00\%. These gains are larger than the English-standard gain, which rises only from 98.40\% to 99.07\%. This suggests that, on lower-difficulty general propositions, increasing model scale is especially effective when the model must first normalize a non-canonical Chinese expression before making the final logical judgment.

Scaling also improves the Difficult aligned set after Qwen3-14B. Difficult English improves strongly with scale: Qwen3-8B scores 80.50\%, Qwen3-14B scores 93.02\%, and Qwen3-32B scores 96.05\%. Difficult Chinese average accuracy rises from 52.77\% for Qwen3-8B to 54.73\% for Qwen3-14B and 82.32\% for Qwen3-32B. Even so, Qwen3-32B still shows a large gap between difficult English and difficult rhetorical Chinese: 96.05\% versus 69.35\%. The result indicates that larger models substantially reduce but do not eliminate Chinese surface-normalization errors.

\subsection{Logical-template-level difficulty}
Figure~\ref{fig:main_results}B and the logical-template-family-level tables in Appendix~\ref{app:detailed} reveal persistent template-level weaknesses. Propositional equivalence is the most consistent bottleneck. Qwen3-8B remains low on natural Chinese (41.11\%) and rhetorical Chinese (30.67\%). Qwen3-14B improves natural Chinese equivalence to 60.22\%, but colloquial Chinese equivalence is still only 34.22\%. Qwen3-32B substantially improves this family, reaching 78.67\% on standard Chinese, 78.00\% on natural Chinese, 90.89\% on colloquial Chinese, 94.44\% on rhetorical Chinese, and 85.78\% on perturbed Chinese. Ministral-8B shows an even sharper English--Chinese contrast: 94.67\% on English equivalence, but 23.56\% on standard Chinese and 8.00\% on rhetorical Chinese. GLM-5.1 is strong on English equivalence (98.22\%) and standard Chinese equivalence (78.22\%), but drops to 34.44\% on natural Chinese and 43.33\% on rhetorical Chinese. These results indicate that equivalence laws require global scope tracking and truth-condition comparison, which Chinese paraphrase easily disrupts.

Invalid inference is also diagnostic. Strong models often perform well when invalid forms are explicit, but fail when Chinese surface realization encourages a plausible but invalid conclusion. For example, Qwen3-8B drops from 99.75\% on English invalid inference to 41.50\% on rhetorical Chinese invalid inference. GLM-5.1 drops from 98.50\% to 90.50\% in the same logical-template family. These failures are not ordinary reasoning failures on valid rules; they are failures to resist pragmatically tempting but logically invalid inferences such as affirming the consequent, denying the antecedent, illicit conversion, and existential fallacy.

\subsection{Language-dependent fallacy detection and targeted English augmentation}

A further observation concerns the fourth logical-template family, common invalid inference and logical fallacies. Prior work such as LogicAsker reports that many LLMs exhibit substantial weaknesses in learning and applying formal logical rules, including common fallacy-related reasoning patterns~\citep{wan2024logicasker}. Our results refine this conclusion from a cross-lingual perspective. The weakness is not always a language-independent inability to recognize fallacies. Instead, fallacy detection is highly sensitive to the surface language and to how explicitly the invalid logical relation is expressed.

In our aligned template results, several models perform better on the standard Chinese translation than on the English standard expression for common invalid inference. This pattern is especially visible in the fourth logical-template family. For example, Ministral-3B obtains 91.50\% on English invalid inference but reaches 100\% on all five Chinese variants. Ministral-8B obtains 73.25\% on English invalid inference but 94.75\% on standard Chinese. GLM-5.1 also improves from 98.50\% on English to 99.50\% on standard Chinese. Although the magnitude differs across models, the direction is notable: the Chinese standard translation is not necessarily harder than English. In some fallacy patterns, it is easier, probably because the Chinese translation makes the invalid inference more explicit or more naturally matches the model's learned Chinese reasoning patterns.

This finding is important because it prevents an over-general interpretation of low fallacy accuracy. If a model fails on an English fallacy instance but succeeds on a logically equivalent Chinese translation, then the failure cannot be attributed only to the abstract fallacy itself. It may instead arise from the English wording, the compactness of the English template, or the lack of explicit explanation of the relation between premises and conclusion. In other words, the model may not be uniformly weak at recognizing a fallacy; it may be weak at recovering the intended logical form from a particular surface realization.

As a targeted diagnostic case study rather than a comprehensive augmentation benchmark, we further examined several special logical forms whose original English versions produced unexpectedly low accuracy. We then created augmented English versions in which the same logical relation was expressed more explicitly. Table~\ref{tab:english_aug_fallacy} reports results for five such patterns on the Qwen3 series. The effect is striking. For branch reasoning, Qwen3-8B improves from 10\% to 100\%, and Qwen3-14B improves from 0\% to 100\%. Similar improvements appear for existential fallacy, illicit major, and illicit minor. 

\begin{table*}[t]
\centering
\scriptsize
\begin{tabular}{llccccc}
\toprule
Model & Setting 
& Erroneous quantifier distribution 
& Existential fallacy 
& Illicit major 
& Illicit minor 
& Branch reasoning \\
\midrule
\multirow{4}{*}{Qwen3-8B}
& Original English      & 62  & 94  & 90  & 92  & 10  \\
& Augmented English     & 100 & 100 & 100 & 100 & 100 \\
& Original translation  & 52  & 98  & 62  & 98  & 52  \\
& Augmented translation & 98  & 100 & 100 & 100 & 100 \\
\midrule
\multirow{4}{*}{Qwen3-14B}
& Original English      & 100 & 100 & 96  & 98  & 0   \\
& Augmented English     & 100 & 100 & 100 & 100 & 100 \\
& Original translation  & 52  & 100 & 100 & 100 & 40  \\
& Augmented translation & 100 & 100 & 100 & 100 & 100 \\
\bottomrule
\end{tabular}
\caption{Effect of explicit English augmentation on five special logical patterns in the Qwen3 series. The augmented English versions preserve the same underlying logical forms but state the relevant logical relation more explicitly. The large gains indicate that some apparent fallacy-detection failures are caused by surface-form recovery rather than by the absence of the underlying reasoning ability.}
\label{tab:english_aug_fallacy}
\end{table*}

The case of denying a conjunct is particularly revealing. This pattern has the form $\neg(P \land Q), \neg P, \therefore Q$, and the conclusion is invalid: from the fact that $P$ and $Q$ cannot both hold, and from the fact that $P$ does not hold, it does not follow that $Q$ must hold. Although this pattern was not included as one of the nine main logical-template families, it exposes the same surface-sensitivity. This again suggests that the English formulation of some fallacy patterns can be unusually difficult for the model, while a Chinese formulation may make the invalidity easier to detect.

Overall, these results qualify the claim that LLMs are simply weak at logical fallacy recognition. Our evidence suggests a more precise conclusion: LLMs are often weak at recovering the intended logical form from compact or under-specified surface expressions, and this weakness interacts strongly with language. When the same fallacy is rendered as a clearer Chinese standard translation, or when the English expression is augmented to state the logical relation more explicitly, performance can increase sharply. Therefore, evaluating fallacy understanding requires controlling not only the abstract logical schema, but also the linguistic realization through which the schema is presented.

\subsection{Chinese-to-English back-translation experiments}

To separate failures of formal inference from failures of Chinese surface-to-logical-form normalization, we evaluate an additional back-translated English condition. For the aligned template benchmark, the standard Chinese realization is translated back into English with the checkpoint documented in Appendix~\ref{app:construction_audit} and then evaluated with the same label set and scoring script. Table~\ref{tab:std_backtranslation} reports the complete family-level results on the General aligned set, and Table~\ref{tab:difficult_backtranslation} reports the same diagnostic on the Difficult aligned set, for Qwen3-8B, Qwen3-14B, Qwen3-32B, Ministral-8B-2512, and GLM-5.1. The back-translated condition is not a new logical task: it contains the same latent templates as the standard Chinese condition. However, the transformation may simplify wording, alter pragmatic force, change negation scope or condition direction, remove Chinese-specific cues, or introduce English-side artifacts; it is therefore treated as a diagnostic rather than a neutral control.

\begin{table*}[t]
\centering
\scriptsize
\setlength{\tabcolsep}{2.1pt}
\resizebox{\textwidth}{!}{%
\begin{tabular}{lrrrrrrrrrrrrrrr}
\toprule
\multirow{2}{*}{Logical-template family} & \multicolumn{3}{c}{Qwen3-8B} & \multicolumn{3}{c}{Qwen3-14B} & \multicolumn{3}{c}{Qwen3-32B} & \multicolumn{3}{c}{Ministral-8B} & \multicolumn{3}{c}{GLM-5.1} \\
\cmidrule(lr){2-4}\cmidrule(lr){5-7}\cmidrule(lr){8-10}\cmidrule(lr){11-13}\cmidrule(lr){14-16}
& En & Ch-Std & BT-En & En & Ch-Std & BT-En & En & Ch-Std & BT-En & En & Ch-Std & BT-En & En & Ch-Std & BT-En \\
\midrule
Basic connectives & 100.00 & 86.29 & 100.00 & 100.00 & 100.00 & 100.00 & 100.00 & 100.00 & 100.00 & 99.71 & 100.00 & 100.00 & 98.86 & 91.14 & 100.00 \\
Propositional equivalence laws & 93.56 & 61.78 & 92.44 & 94.67 & 60.22 & 94.44 & 99.33 & 78.67 & 99.11 & 94.67 & 23.56 & 95.33 & 98.22 & 78.22 & 98.67 \\
Propositional inference rules & 100.00 & 98.36 & 100.00 & 100.00 & 100.00 & 100.00 & 100.00 & 100.00 & 100.00 & 99.45 & 87.27 & 99.45 & 100.00 & 98.73 & 100.00 \\
Common invalid inferences / fallacies & 99.75 & 98.00 & 99.50 & 100.00 & 99.25 & 100.00 & 93.75 & 92.00 & 95.75 & 73.25 & 94.75 & 73.25 & 98.50 & 99.50 & 98.25 \\
Predicate-logic basics & 99.60 & 92.40 & 99.60 & 99.20 & 99.20 & 96.80 & 100.00 & 100.00 & 100.00 & 98.80 & 84.80 & 100.00 & 94.00 & 90.00 & 88.00 \\
Quantifier equivalence laws & 96.80 & 97.60 & 94.40 & 100.00 & 96.80 & 100.00 & 100.00 & 100.00 & 100.00 & 80.40 & 48.80 & 78.00 & 94.80 & 79.20 & 91.20 \\
Predicate-logic inference rules & 100.00 & 100.00 & 100.00 & 100.00 & 100.00 & 100.00 & 100.00 & 100.00 & 100.00 & 100.00 & 100.00 & 100.00 & 100.00 & 100.00 & 100.00 \\
Multi-step reasoning & 97.50 & 93.50 & 97.50 & 100.00 & 100.00 & 100.00 & 100.00 & 100.00 & 100.00 & 96.00 & 100.00 & 97.50 & 99.50 & 97.00 & 99.50 \\
Relation logic & 99.00 & 93.50 & 99.50 & 100.00 & 100.00 & 100.00 & 100.00 & 100.00 & 100.00 & 78.00 & 98.00 & 75.50 & 98.00 & 99.50 & 99.00 \\
\midrule
General aligned set weighted average & 98.40 & 90.53 & 99.10 & 99.13 & 93.60 & 98.90 & 99.07 & 95.73 & 99.30 & 92.03 & 79.83 & 91.20 & 98.30 & 92.60 & 97.73 \\
\bottomrule
\end{tabular}%
}
\caption{Accuracy (\%) for the standard Chinese back-translation probe on the General aligned set. En is the original English standard expression, Ch-Std is the original standard Chinese translation, and BT-En is the LLM generated English back-translation of standard Chinese translation.}
\label{tab:std_backtranslation}
\end{table*}

The pattern in Table~\ref{tab:std_backtranslation} is especially clear on the 3,000-item General aligned set: back-translation often brings performance close to the original English-reference level. The aggregate row is computed as a weighted average over all General aligned set logical expressions rather than as a simple mean over the nine logical-template families, because the families contain different numbers of base expressions. Under this weighting, every main model scores substantially higher on BT-En than on standard Chinese: Qwen3-8B rises from 90.53\% to 99.10\%, Qwen3-14B from 93.60\% to 98.90\%, Qwen3-32B from 95.73\% to 99.30\%, Ministral-8B from 79.83\% to 91.20\%, and GLM-5.1 from 92.60\% to 97.73\%. These BT-En scores are also close to the corresponding English-reference scores: 98.40\%, 99.13\%, 99.07\%, 92.03\%, and 98.30\%, respectively. Thus, the same logical content becomes much easier for the models once the standard Chinese surface realization is converted back into English. The largest gains appear in families where Chinese wording changes the recoverable logical form: propositional equivalence laws, predicate basics, quantifier equivalence laws, and relation logic. This result strengthens the English--Chinese contrast but does not establish a single causal explanation. It is consistent with difficulty in normalizing Chinese expressions into the intended logical representation, while translation-induced simplification and English-side artifacts may also contribute. Table~\ref{tab:difficult_backtranslation} extends the same diagnostic to the 2,000-item Difficult aligned set.

\begin{table*}[t]
\centering
\scriptsize
\setlength{\tabcolsep}{4pt}
\begin{tabular}{lrrrr}
\toprule
Model & En & Ch-Std & BT-En & $\Delta$BT--Ch \\
\midrule
Qwen3-8B & 80.50 & 52.50 & 74.60 & +22.10 \\
Qwen3-14B & 93.02 & 58.28 & 81.75 & +23.47 \\
Qwen3-32B & 96.05 & 83.10 & 79.00 & -4.10 \\
Ministral-8B & 68.55 & 64.55 & 67.65 & +3.10 \\
GLM-5.1 & 84.70 & 81.00 & 62.95 & -18.05 \\
\bottomrule
\end{tabular}
\caption{Accuracy (\%) for the standard Chinese back-translation probe on the Difficult aligned set. En is the original English condition, Ch-Std is the original standard Chinese condition, and BT-En is the English back-translation of the standard Chinese field. $\Delta$BT--Ch reports BT-En minus Ch-Std.}
\label{tab:difficult_backtranslation}
\end{table*}

On the Difficult aligned set, back-translation is more mixed than on the General aligned set. Qwen3-8B and Qwen3-14B recover substantially after back-translation, improving by 22.10 and 23.47 points over standard Chinese. Ministral-8B shows a smaller gain of 3.10 points. In contrast, Qwen3-32B drops from 83.10\% on standard Chinese to 79.00\% on BT-En, and GLM-5.1 drops from 81.00\% to 62.95\%. This suggests that back-translation is a useful diagnostic but not a uniformly simplifying transformation: for difficult logical problems, the translated English surface realization can sometimes remove helpful Chinese cues, introduce new ambiguity, or interact differently with the model's decision boundary.

We next apply the same diagnostic idea to the 15 statements specific to Chinese context. The original suite of statements specific to Chinese context contains 15 phenomenon types, with 100 Chinese examples per type, for a total of 1,500 Chinese-only samples. We translate these samples back into English and evaluate the same five main models. Tables~\ref{tab:phenomena_chinese} and~\ref{tab:phenomena_backtranslation} report the complete phenomenon-level results for the original Chinese inputs and their back-translated English counterparts.

\begin{table*}[t]
\centering
\scriptsize
\setlength{\tabcolsep}{4pt}
\resizebox{\textwidth}{!}{%
\begin{tabular}{lrrrrr}
\toprule
Phenomenon & Qwen3-8B & Qwen3-14B & Qwen3-32B & Ministral-8B & GLM-5.1 \\
\midrule
Conditional particles: as long as /  only if / unless / otherwise & 78 & 69 & 60 & 60 & 60 \\
Not all vs. all not & 81 & 71 & 51 & 43 & 69 \\
Not necessarily vs. necessarily not & 39 & 43 & 40 & 40 & 50 \\
Double and multiple negation & 69 & 59 & 40 & 40 & 57 \\
Rhetorical questions & 59 & 40 & 61 & 44 & 63 \\
Ellipsis of subject, object, condition, or conclusion & 75 & 65 & 20 & 20 & 89 \\
Polysemy and homographs & 47 & 78 & 93 & 100 & 99 \\
Segmentation ambiguity & 85 & 86 & 40 & 51 & 100 \\
Fuzzy quantifiers: all/some/most/basically/not necessarily & 88 & 100 & 10 & 12 & 91 \\
Comparative structures & 78 & 100 & 40 & 46 & 94 \\
Temporal order vs. causality & 13 & 40 & 20 & 0 & 18 \\
Concession, contrast, and embedded conditions & 80 & 85 & 78 & 64 & 84 \\
Irony & 48 & 79 & 96 & 100 & 95 \\
Legal, rule-style, and notice-style Chinese & 81 & 87 & 80 & 78 & 75 \\
Brand names, abbreviations, and homophone-based concept shifts & 87 & 94 & 100 & 98 & 100 \\
\midrule
Average & 67.2 & 70.07 & 55.27 & 53.07 & 76.27 \\
\bottomrule
\end{tabular}%
}
\caption{Accuracy (\%) on the original Chinese version of the 15 statements specific to Chinese context. Each phenomenon contains 100 generated Chinese examples.}
\label{tab:phenomena_chinese}
\end{table*}

\begin{table*}[t]
\centering
\scriptsize
\setlength{\tabcolsep}{4pt}
\resizebox{\textwidth}{!}{%
\begin{tabular}{lrrrrr}
\toprule
Phenomenon & Qwen3-8B & Qwen3-14B & Qwen3-32B & Ministral-8B & GLM-5.1 \\
\midrule
Conditional particles: as long as /  only if / unless / otherwise & 62 & 62 & 60 & 68 & 78 \\
Not all vs. all not & 93 & 93 & 60 & 99 & 100 \\
Not necessarily vs. necessarily not & 28 & 40 & 40 & 48 & 85 \\
Double and multiple negation & 75 & 40 & 40 & 53 & 62 \\
Rhetorical questions & 74 & 55 & 64 & 83 & 82 \\
Ellipsis of subject, object, condition, or conclusion & 98 & 61 & 30 & 100 & 100 \\
Polysemy and homographs & 100 & 100 & 100 & 92 & 100 \\
Segmentation ambiguity & 74 & 83 & 52 & 100 & 100 \\
Fuzzy quantifiers: all/some/most/basically/not necessarily & 53 & 100 & 10 & 84 & 100 \\
Comparative structures & 89 & 57 & 40 & 80 & 91 \\
Temporal order vs. causality & 40 & 34 & 20 & 94 & 67 \\
Concession, contrast, and embedded conditions & 83 & 79 & 80 & 70 & 84 \\
Irony & 57 & 18 & 96 & 29 & 93 \\
Legal, rule-style, and notice-style Chinese & 94 & 81 & 80 & 82 & 93 \\
Brand names, abbreviations, and homophone-based concept shifts & 70 & 91 & 100 & 30 & 96 \\
\midrule
Average & 72.67 & 66.27 & 58.13 & 74.13 & 88.73 \\
\bottomrule
\end{tabular}%
}
\caption{Accuracy (\%) on the English back-translated version of the 15 statements specific to Chinese context. The same examples are translated from Chinese into English and evaluated with the same prompt and label parser.}
\label{tab:phenomena_backtranslation}
\end{table*}

The phenomenon-level back-translation results show that the English conversion often improves accuracy, but the effect is not uniform. Qwen3-8B improves on average, from 67.20\% to 72.67\%. Qwen3-32B also improves slightly, from 55.27\% to 58.13\%. Ministral-8B improves substantially, from 53.07\% to 74.13\%, and GLM-5.1 improves from 76.27\% to 88.73\%. Qwen3-14B is the exception among the Qwen models: it drops from 70.07\% on the original Chinese phenomenon set to 66.27\% on the back-translated English version. This exception matters because it shows that back-translation is not simply easier for every model; translation can also erase Chinese cues that a particular model used successfully or introduce English formulations that interact badly with its decision boundary.

The phenomenon-level results identify where back-translation helps most. For Ministral-8B, ellipsis rises from 20\% to 100\%, temporal-order versus causality rises from 0\% to 94\%, and fuzzy quantifiers rise from 12\% to 84\%. For GLM-5.1, not-all versus all-not rises from 69\% to 100\%, fuzzy quantifiers rise from 91\% to 100\%, and ellipsis rises from 89\% to 100\%. Qwen3-8B also benefits on ellipsis, legal or rule-style Chinese, and not-all versus all-not. At the same time, back-translation can hurt phenomenon types such as irony, brand-name or homophone-based concept shifts, and some comparative structures. These phenomenon types depend on pragmatic, lexical, or culturally grounded cues that may be weakened by translation. Together with Table~\ref{tab:std_backtranslation}, these results support a two-stage interpretation: models often have usable reasoning ability once the structure is made explicit, but robust Chinese logical understanding requires a separate and still fragile normalization step.

\subsection{Degenerate and partially degenerate behavior}

Gold-label frequencies are reported in Table~\ref{tab:label_distribution}; they are essential for interpreting biased output strategies. Appendix~\ref{app:bias_metrics} reports bias-aware metrics and prediction distributions for the General aligned set.

Qwen3-0.6B exhibits the clearest output degeneration. Its logical-template-family tables show repeated identical scores across surface-realization conditions, while its accuracy on invalid-inference families collapses to 0\%. This pattern is consistent with a near-universal \yes{} output. In the General aligned set, \yes{} accounts for 78.33\% of the gold labels, compared with 21.67\% for \no{} and 0\% for \unk{}; in the Difficult aligned set, the corresponding proportions are 55.00\%, 45.00\%, and 0\%. Consequently, an almost-always-\yes{} strategy can obtain deceptively high aggregate accuracy, especially on the General aligned set, while failing families whose correct answers are predominantly \no{} or \unk{}. Qwen3-0.6B should therefore be interpreted as a task-following and calibration failure case rather than as a small but competent logical reasoner.

Ministral-3B shows a different but related form of degeneration. Rather than predicting \yes{} almost universally, it appears strongly biased toward \no{}. The clearest evidence comes from the invalid-inference family: on the General aligned set, it achieves 100\% accuracy on all five Chinese variants and 91.50\% on English, because invalid inferences are labeled \no{} under our task definition. At the same time, it performs extremely poorly on many predominantly \yes{} families. For example, on propositional equivalence it scores 19.11\% in English and nearly 0\% on several Chinese forms; on predicate basics it scores only 6.00\% on standard Chinese and 0\% on rhetorical Chinese; and on basic connectives it scores only 18\% on rhetorical Chinese. The label distribution further helps explain why its Difficult aligned set can appear stronger than its General aligned set: \no{} accounts for 45.00\% of the Difficult aligned set but only 21.67\% of the General aligned set. Thus, a strong \no{} bias can inflate aggregate accuracy on the Difficult aligned set without indicating more reliable logical reasoning.

\section{Forty Difficult Logical Problems}
\label{app:difficult_skeletons}
This appendix lists the 40 difficult logical skeletons used in the Difficult aligned set. These skeletons are adapted from and inspired by standard mathematical-logic textbook exercises on first-order semantics, quantifier reasoning, relational properties, uniqueness, equality, and countermodel construction~\citep{enderton2001mathematical,mendelson2015introduction}. The formulas are not copied verbatim from a single textbook; instead, they are normalized into a unified first-order formula format with binary formula-level labels under standard first-order semantics with a non-empty domain.

\clearpage
\onecolumn

\begin{sloppypar}
\scriptsize

\setlength{\LTleft}{0pt}
\setlength{\LTright}{0pt}
\renewcommand{\arraystretch}{1.08}

\begin{longtable}{
    @{}
    >{\raggedright\arraybackslash}p{0.040\textwidth}
    @{\hspace{3pt}}
    >{\raggedright\arraybackslash}p{0.865\textwidth}
    @{\hspace{3pt}}
    >{\centering\arraybackslash}p{0.055\textwidth}
    @{}
}
\caption{The 40 difficult logical problems from textbooks.}
\label{tab:app_difficult_40}\\

\toprule
ID & Logical expression & Label \\
\midrule
\endfirsthead

\toprule
ID & Logical expression & Label \\
\midrule
\endhead

\midrule
\multicolumn{3}{r}{\scriptsize Continued on the next page}
\\
\endfoot

\bottomrule
\endlastfoot

D1 & $\exists x\forall y(P(x,y)\leftrightarrow \neg P(y,x))$ & False \\
D2 & $\forall x(\exists yP(x,y)\leftrightarrow \forall z\neg P(z,x))$ & False \\
D3 & $\forall x(P(x)\rightarrow \exists y(P(y)\land \forall z(P(z)\rightarrow (R(x,z)\rightarrow R(y,z)))))$ & True \\
D4 & $\forall x\exists y\forall z((P(x,y)\land Q(z))\rightarrow R(x,z))\rightarrow \exists u\forall x\exists zR(x,z)$ & False \\
D5 & $\forall x\exists y(P(x,y)\land \forall z(P(x,z)\rightarrow y=z))\rightarrow \exists x\forall y\neg P(y,x)$ & False \\
D6 & $(\forall x\exists y\forall z((P(x,y)\land P(y,z))\rightarrow P(x,z))\land \forall x\neg P(x,x))\rightarrow \neg\exists x\exists yP(x,y)$ & False \\
D7 & $(\exists x\forall y(P(x,y)\leftrightarrow \forall z(P(y,z)\rightarrow P(x,z)))\land \exists yP(y,y))\rightarrow \exists xP(x,x)$ & True \\
D8 & $(\forall x(P(x)\leftrightarrow \forall y(P(y)\leftrightarrow R(x,y)))\land \forall x\forall y(R(x,y)\leftrightarrow R(y,x)))\rightarrow \exists xP(x)$ & False \\
D9 & $(\forall x\forall y\forall z((R(x,y)\land R(y,z))\rightarrow R(x,z))\land \forall x\neg R(x,x))\rightarrow \neg\exists x\exists yR(x,y)$ & False \\
D10 & $\forall x\exists y(P(x,y)\leftrightarrow \forall zP(z,x))\rightarrow \exists u\forall x(P(x,u)\leftrightarrow \forall zP(z,x))$ & False \\
D11 & $\exists x\forall y(P(x)\rightarrow Q(y))\leftrightarrow \forall y\exists x(P(x)\rightarrow Q(y))$ & True \\
D12 & $\forall x\exists y(P(x,y)\land \forall z(Q(x,z)\rightarrow R(y,z)))\rightarrow \exists y\forall x\forall z(P(x,y)\land (Q(x,z)\rightarrow R(y,z)))$ & False \\
D13 & $(\exists x\forall y(P(y)\leftrightarrow R(x,y))\land \forall x\forall y(R(y,x)\rightarrow R(y,y)))\rightarrow \forall xP(x)$ & False \\
D14 & $(\forall x\forall y(R(x,y)\rightarrow \exists z(P(x,z)\land P(y,z)))\land \forall x\exists yR(x,y))\rightarrow \forall x\exists zP(x,z)$ & True \\
D15 & $\forall x\exists y\forall z(P(x,z)\leftrightarrow z=y)\rightarrow \forall x\exists yP(x,y)$ & True \\
D16 & $\forall x\exists yP(x,y)\rightarrow \exists y\forall xP(x,y)$ & False \\
D17 & $\exists y\forall xP(x,y)\rightarrow \forall x\exists yP(x,y)$ & True \\
D18 & $\forall x(P(x)\rightarrow Q(x))\rightarrow (\exists xP(x)\rightarrow \exists xQ(x))$ & True \\
D19 & $(\forall x(P(x)\rightarrow Q(x))\land \exists x\neg Q(x))\rightarrow \exists x\neg P(x)$ & True \\
D20 & $(\forall x(P(x)\rightarrow Q(x))\land \exists xQ(x))\rightarrow \exists xP(x)$ & False \\
D21 & $\neg\exists x(P(x)\land Q(x))\leftrightarrow \forall x(P(x)\rightarrow \neg Q(x))$ & True \\
D22 & $\neg\forall xP(x)\rightarrow \forall x\neg P(x)$ & False \\
D23 & $\forall x(P(x)\lor Q(x))\rightarrow (\forall xP(x)\lor \forall xQ(x))$ & False \\
D24 & $(\forall xP(x)\lor \forall xQ(x))\rightarrow \forall x(P(x)\lor Q(x))$ & True \\
D25 & $\exists x(P(x)\land Q(x))\rightarrow (\exists xP(x)\land \exists xQ(x))$ & True \\
D26 & $(\exists xP(x)\land \exists xQ(x))\rightarrow \exists x(P(x)\land Q(x))$ & False \\
D27 & $(\forall x\forall y(R(x,y)\rightarrow R(y,x))\land \forall x\forall y\forall z((R(x,y)\land R(y,z))\rightarrow R(x,z))\land \exists x\exists yR(x,y))\rightarrow \exists xR(x,x)$ & True \\
D28 & $(\forall x\forall y(R(x,y)\rightarrow R(y,x))\land \exists x\forall yR(x,y))\rightarrow \forall x\exists yR(x,y)$ & True \\
D29 & $(\forall x\forall y(R(x,y)\rightarrow R(y,x))\land \forall x\exists yR(x,y))\rightarrow \forall xR(x,x)$ & False \\
D30 & $(\forall x\forall y((R(x,y)\land R(y,x))\rightarrow x=y)\land R(a,b)\land R(b,a))\rightarrow a=b$ & True \\
D31 & $(\forall x\forall y(R(x,y)\rightarrow R(y,x))\land \forall x\forall y((R(x,y)\land R(y,x))\rightarrow x=y))\rightarrow \forall x\forall y(R(x,y)\rightarrow x=y)$ & True \\
D32 & $(\forall x\forall y(R(x,y)\leftrightarrow R(y,x))\land \forall x\forall y\forall z((R(x,y)\land R(y,z))\rightarrow R(x,z)))\rightarrow \forall x(R(x,x)\lor \forall y\neg R(x,y))$ & True \\
D33 & $\forall x\forall y(R(x,y)\rightarrow R(y,x))\rightarrow \forall x\forall y\forall z((R(x,y)\land R(y,z))\rightarrow R(x,z))$ & False \\
D34 & $(\forall x\forall y(R(x,y)\leftrightarrow R(y,x))\land \forall x\forall y(R(x,y)\leftrightarrow S(x,y)))\rightarrow \forall x\forall y(S(x,y)\leftrightarrow S(y,x))$ & True \\
D35 & $(\forall x\exists yR(x,y)\land \forall x\forall y\forall z((R(x,y)\land R(x,z))\rightarrow y=z))\rightarrow \exists y\forall xR(x,y)$ & False \\
D36 & $(\forall x\exists yR(x,y)\land \forall x\forall y(R(x,y)\rightarrow P(y)))\rightarrow \forall x\exists yP(y)$ & True \\
D37 & $\forall x\exists y(R(x,y)\land P(y))\rightarrow \exists yP(y)$ & True \\
D38 & $(\exists x\forall y(R(x,y)\leftrightarrow P(y))\land \forall y(P(y)\rightarrow Q(y)))\rightarrow \exists x\forall y(R(x,y)\rightarrow Q(y))$ & True \\
D39 & $(\forall x(P(x)\rightarrow Q(x))\land \forall x(Q(x)\rightarrow R(x)))\rightarrow \forall x(P(x)\rightarrow R(x))$ & True \\
D40 & $(\exists x(P(x)\land \forall y(P(y)\rightarrow y=x))\land \exists xP(x))\rightarrow \forall x\forall y((P(x)\land P(y))\rightarrow x=y)$ & True \\

\bottomrule
\end{longtable}
\clearpage
\twocolumn
\end{sloppypar}
\section{Examples of Statements Specific to Chinese Context}
\label{app:cn_examples}

\newenvironment{cnquote}
{\begin{quote}\footnotesize}
{\end{quote}}

Chinese examples below are presented in the order \textbf{Chinese--Pinyin--English}. Pinyin follows tone-marked Hanyu Pinyin; proper names are capitalized, and repeated key markers are not re-romanized elsewhere after their first informative occurrence.

\subsection{Romanization of the Figure~\ref{fig:pipeline} Example}
\label{app:figure_romanization}

The five Chinese strings shown in Figure~\ref{fig:pipeline} are aligned with the English reference and retain the same intended logical label.

\begin{cnquote}
\textbf{Ch-Std Chinese:} \begin{CJK*}{UTF8}{gbsn}如果李明认真阅读了题目，那么他考试通过了。李明考试通过了。因此，李明认真阅读了题目。\end{CJK*}\\
\textbf{English:} If Li Ming read the question carefully, then he passed the exam. Li Ming passed the exam. Therefore, Li Ming read the question carefully.
\end{cnquote}

\begin{cnquote}
\textbf{Ch-Nat Chinese:} \begin{CJK*}{UTF8}{gbsn}如果李明认真阅读了题目，他就会通过考试。他通过了考试。所以，李明认真阅读了题目。\end{CJK*}\\
\textbf{English:} If Li Ming read the question carefully, he would pass the exam. He passed the exam. Therefore, Li Ming read the question carefully.
\end{cnquote}

\begin{cnquote}
\textbf{Ch-Col Chinese:} \begin{CJK*}{UTF8}{gbsn}李明要是把题认真阅读了，考试就能过。他考过了。那说明他认真阅读过题目。\end{CJK*}\\
\textbf{English:} If Li Ming read the question carefully, he could pass the exam. He passed it. That shows that he read the question carefully.
\end{cnquote}

\begin{cnquote}
\textbf{Ch-Rhet Chinese:} \begin{CJK*}{UTF8}{gbsn}李明都考试通过了，难道还不能说明他认真阅读过题目吗？\end{CJK*}\\
\textbf{English:} Li Ming passed the exam; does that not show that he read the question carefully?
\end{cnquote}

\begin{cnquote}
\textbf{Ch-Pert Chinese:} \begin{CJK*}{UTF8}{gbsn}已知只要李明认真阅读题目，他就会通过考试；而他既然考试通过了。那不能推出他之前一定认真阅读了题目吗？\end{CJK*}\\
\textbf{English:} Given that Li Ming would pass the exam if he read the question carefully, and that he passed, can we not conclude that he must have read the question carefully beforehand?
\end{cnquote}

\subsection{Representative Chinese-Only Examples}

This subsection provides representative examples for the 3 Chinese-context phenomenon types summarized in Table~\ref{tab:cn_phenomena}. They illustrate how surface-level expressions can obscure recovery of the underlying logical form.

\begin{enumerate}
    \item \textbf{Condition markers: as long as, only if, unless, otherwise.}
    
    This type tests whether a model distinguishes sufficient conditions from necessary conditions.
    
    \begin{cnquote}
    \textbf{Chinese:} \begin{CJK*}{UTF8}{gbsn}只有报名成功，才能参加考试。小王参加了考试。\end{CJK*}\\
    \textbf{English:} Only if registration is successful can one take the exam. Xiao Wang took the exam.\\
    \textbf{Question:} Did Xiao Wang successfully register?\\
    \textbf{Answer:} Yes.
    \end{cnquote}
    
    \begin{cnquote}
    \begin{CJK*}{UTF8}{gbsn}\textbf{Chinese:} 只要报名成功，就能参加考试。小王参加了考试。\end{CJK*}\\
    \textbf{English:} If registration is successful, then one can take the exam. Xiao Wang took the exam.\\
    \textbf{Question:} Did Xiao Wang successfully register?\\
    \textbf{Answer:} Not necessarily.
    \end{cnquote}

    
    

    \item \textbf{``Not necessarily'' vs. ``necessarily not''.}
    
    This type tests whether a model avoids confusing uncertainty with logical negation.
    
    \begin{cnquote}
    \textbf{Chinese:} \begin{CJK*}{UTF8}{gbsn}小王不一定会通过审核。\end{CJK*}\\
    \textbf{English:} It is not certain that Xiao Wang will pass the review.\\
    \textbf{Question:} Does this mean Xiao Wang will not pass the review?\\
    \textbf{Answer:} Not necessarily.
    \end{cnquote}

\end{enumerate}

\section{Generation, Verification, and Evaluation Prompts}
\label{app:prompts}

This appendix provides representative prompts used in the construction and evaluation workflow. The prompts are templates: bracketed fields such as \texttt{[LOGICAL TEMPLATE]} and \texttt{[QUERY TEXT]} are replaced with item-specific content.

\subsection{Surface Realization Prompt}
\begin{quote}
\small
\setlength{\parskip}{2pt}
\setlength{\parindent}{0pt}
\ttfamily
You are helping construct a controlled logical-reasoning benchmark.

Input fields: logical template, logical family, premises, hypothesis or question, and label.

Generate one English standard expression and five Chinese surface realizations. The English standard expression should state the premises and the target question in a clear, explicit, and logically regular form.

The five Chinese surface realizations should follow these definitions:
\begin{enumerate}[leftmargin=*,nosep]
    \item Standard Chinese: a faithful Chinese rendering that closely follows the logical structure of the English standard expression.
    \item Natural written Chinese: a fluent written Chinese version that sounds natural in ordinary written discourse while preserving the same logic.
    \item Colloquial Chinese: an everyday spoken-style Chinese version, allowing informal wording, discourse particles, or mild ellipsis, but not changing the logical content.
    \item Rhetorical-question Chinese: a Chinese version that expresses the same target judgment through a rhetorical-question form; the rhetorical form may change the discourse style but must not change the intended label.
    \item Perturbed Chinese: a Chinese version that preserves the same intended logical relation and label while adding controlled distractors, ambiguity, word-order variation, or pragmatic interference.
\end{enumerate}

Constraints: preserve the same premises, the same target question, the same latent logical relation, and the same label. Do not add a premise that changes the answer. Do not remove information needed for the judgment. The rhetorical version may change the discourse form, and the perturbed version may add distractors or ambiguity, but neither may change the intended label.

Return the six realizations in separate fields.
\end{quote}

\subsection{Prompt for Statements Specific to Chinese Context}
\begin{quote}\small\ttfamily
You are helping construct the Chinese-context part of a controlled logical-reasoning benchmark. This part targets the 15 phenomenon types that are specific to Chinese context, such as conditional markers, negation scope, rhetorical questions, ellipsis, polysemy, homophones, punctuation ambiguity, vague quantifiers, comparative structures, causal order, concession and contrast, irony, rule-style Chinese, and concept shifts caused by brand names, abbreviations, or sound-alike expressions.

Given one target Chinese-context phenomenon type, generate Chinese examples that test this specific phenomenon. Each example should contain a complete input text, a question, and a label from yes, no, and unknown. The label must be determined by the information explicitly provided in the input text. The example should make the target Chinese-context phenomenon clear, while still requiring logical judgment rather than simple keyword matching.

If the target type involves irony, polysemy, homophones, ellipsis, rule-style language, or other pragmatic or lexical effects, make the context sufficient for the intended interpretation. Avoid examples whose answer depends on private world knowledge, unstated assumptions, or information outside the input. Do not include an explanation or rationale in the final output.
\end{quote}

\subsection{Verification Prompt}
\begin{quote}\small\ttfamily
Given the English standard expression and the five Chinese surface realizations, check whether all versions express the same premises, ask the same question, and preserve the same label. Identify any version that adds a new premise, removes necessary information, changes the target hypothesis, changes the answer, or introduces uncontrolled ambiguity. Return PASS only if all six realizations preserve the same latent logical relation and label; otherwise return FAIL with reasons.
\end{quote}

\subsection{Evaluation Prompt}
\begin{quote}\small\ttfamily
You are solving a symbolic logic reasoning task.

Read the following facts and question carefully. Determine whether the
conclusion in the question necessarily follows from the facts.

You must answer with exactly one word from the following three options:
yes
no
unknown

Meaning:
- yes: the conclusion necessarily follows.
- no: the conclusion is false or the proposed inference is invalid.
- unknown: the facts are insufficient to determine the conclusion.

Do not output any explanation, punctuation, or extra text.

\end{quote}

\section{Detailed Experimental Tables}
\label{app:detailed}

This appendix reports all experimental values in a more compact but still inspectable organization. The two overview tables are followed by one family-level summary table for each model. Fine-grained results are then regrouped by logical-template family: each table compares all seven models, and each cell reports the six conditions in the fixed order English / Ch-Std / Ch-Nat / Ch-Col / Ch-Rhet / Ch-Pert. This preserves the complete reported results while avoiding a separate table for every model--family pair. Qwen3-0.6B should be read as an output-degeneration diagnostic because it tends to answer \yes{} across almost all inputs. Ministral-3B should also be interpreted carefully because it shows label-asymmetric behavior.

\begin{table*}[t]
\centering
\scriptsize
\setlength{\tabcolsep}{3.5pt}
\begin{tabular}{llrrrrrr}
\toprule
Model & Aligned set & English & Ch-Std & Ch-Nat & Ch-Col & Ch-Rhet & Ch-Pert \\
\midrule
Qwen3-0.6B & General & 78.3 & 78.3 & 78.3 & 78.3 & 78.3 & 78.3 \\
Qwen3-0.6B & Difficult & 55 & 55 & 55 & 55 & 55 & 55 \\
\midrule
Qwen3-8B & General & 98.4 & 90.53 & 80.13 & 80.72 & 68.69 & 77.99 \\
Qwen3-8B & Difficult & 80.5 & 52.5 & 61.58 & 59.48 & 34 & 56.28 \\
Qwen3-14B & General & 99.13 & 93.6 & 86.16 & 85.75 & 75.86 & 85.02 \\
Qwen3-14B & Difficult & 93.02 & 58.28 & 60.42 & 63.12 & 40.22 & 51.62 \\
Qwen3-32B & General & 99.07 & 95.73 & 93.33 & 95.7 & 93.53 & 97 \\
Qwen3-32B & Difficult & 96.05 & 83.1 & 87.8 & 86.15 & 69.35 & 85.2 \\
\midrule
Ministral-3B & General & 59.27 & 41.68 & 40.88 & 30.02 & 31.82 & 37.17 \\
Ministral-3B & Difficult & 73.3 & 47.9 & 49.02 & 49.06 & 46.6 & 46.3 \\
Ministral-8B & General & 92.03 & 79.83 & 82.35 & 79.23 & 70.57 & 84.2 \\
Ministral-8B & Difficult & 68.55 & 64.55 & 77.05 & 72.4 & 49.25 & 72.55 \\
\midrule
GLM-5.1 & General & 98.3 & 92.6 & 84.56 & 84.94 & 78.89 & 83.17 \\
GLM-5.1 & Difficult & 84.7 & 81 & 82.5 & 78.7 & 52.3 & 78.05 \\
\bottomrule
\end{tabular}
\caption{Overall accuracy (\%). The General and Difficult aligned sets use binary labels, whereas the Chinese-only set uses three labels. Ch-Std, Ch-Nat, Ch-Col, Ch-Rhet, and Ch-Pert denote standard Chinese, natural written Chinese, colloquial Chinese, rhetorical-question Chinese, and perturbed Chinese. Qwen3-0.6B and Ministral-3B are marked in the analysis as degenerate or partially degenerate cases because their scores are strongly shaped by output-label bias rather than stable logical judgment.}
\label{tab:app_overall_all}
\end{table*}

\begin{table*}[t]
\centering
\scriptsize
\setlength{\tabcolsep}{4pt}
\resizebox{\textwidth}{!}{%
\begin{tabular}{lrrrrrrr}
\toprule
Phenomenon  & Qwen3-32B & Qwen3-8B & Qwen3-14B & Ministral-8B & Ministral-3B & GLM-5.1 \\
\midrule
Conditional particles: as long as / only if / unless / otherwise & 60 & 78 & 69 & 60 & 18 & 60 \\
Not all vs. all not  & 51 & 81 & 71 & 43 & 0 & 69 \\
Not necessarily vs. necessarily not & 40 & 39 & 43 & 40 & 40 & 50 \\
Double and multiple negation & 40 & 69 & 59 & 40 & 3 & 57 \\
Rhetorical questions  & 61 & 59 & 40 & 44 & 17 & 63 \\
Ellipsis of subject, object, condition, or conclusion & 20 & 75 & 65 & 20 & 0 & 89 \\
Polysemy and homographs & 93 & 47 & 78 & 100 & 100 & 99 \\
Segmentation ambiguity & 40 & 85 & 86 & 51 & 40 & 100 \\
Fuzzy quantifiers: all/some/most/basically/not necessarily & 10 & 88 & 100 & 12 & 0 & 91 \\
Comparative structures & 40 & 78 & 100 & 46 & 1 & 94 \\
Temporal order vs. causality & 20 & 13 & 40 & 0 & 0 & 18 \\
Concession, contrast, and embedded conditions & 78 & 80 & 85 & 64 & 41 & 84 \\
Irony  & 96 & 48 & 79 & 100 & 100 & 95 \\
Legal, rule-style, and notice-style Chinese & 80 & 81 & 87 & 78 & 54 & 75 \\
Brand names, abbreviations, and homophone-based concept shifts  & 100 & 87 & 94 & 98 & 80 & 100 \\
\midrule
Average & 55.27 & 67.2 & 70.07 & 53.07 & 32.93 & 76.27 \\
\bottomrule
\end{tabular}%
}
\caption{Accuracy (\%) on the 15 statements specific to Chinese context. Qwen3-0.6B is shown as unavailable because no separate phenomenon-level table was provided for it; its aggregate output-degeneration behavior is analyzed separately.}
\label{tab:app_cn_phenomena_results}
\end{table*}

\subsection{Logical-template-family summaries by model}
The following seven tables retain the full six-condition family-level summaries for each evaluated model.

\begin{table*}[t]
\centering
\scriptsize
\setlength{\tabcolsep}{3.5pt}
\begin{tabular}{lrrrrrr}
\toprule
  & English & Ch-Std & Ch-Nat & Ch-Col & Ch-Rhet & Ch-Pert \\
\midrule
Basic connectives & 85.71 & 85.71 & 85.71 & 85.71 & 85.71 & 85.71 \\
Prop. equivalence & 100 & 100 & 100 & 100 & 100 & 100 \\
Prop. inference & 100 & 100 & 100 & 100 & 100 & 100 \\
Invalid inference & 0 & 0 & 0 & 0 & 0 & 0 \\
Predicate basics & 100 & 100 & 100 & 100 & 100 & 100 \\
Quantifier equivalence & 80 & 80 & 80 & 80 & 80 & 80 \\
Predicate inference & 87.51 & 87.51 & 87.51 & 87.51 & 85.71 & 85.71 \\
Multi-step reasoning & 75 & 75 & 75 & 75 & 75 & 75 \\
Relation logic & 75 & 75 & 75 & 75 & 75 & 75 \\
General (3,000) & 78.3 & 78.3 & 78.3 & 78.3 & 78.3 & 78.3 \\
Difficult (2,000) & 55 & 55 & 55 & 55 & 55 & 55 \\
\bottomrule
\end{tabular}
\caption{Logical-template-family-level accuracy for Qwen3-0.6B.}
\label{tab:app_qwen306b_summary}
\end{table*}

\begin{table*}[t]
\centering
\scriptsize
\setlength{\tabcolsep}{3.5pt}
\begin{tabular}{lrrrrrr}
\toprule
  & English & Ch-Std & Ch-Nat & Ch-Col & Ch-Rhet & Ch-Pert \\
\midrule
Basic connectives & 100 & 86.29 & 98.29 & 100 & 95.43 & 83.14 \\
Prop. equivalence & 93.56 & 61.78 & 41.11 & 54.22 & 30.67 & 58.44 \\
Prop. inference & 100 & 98.36 & 95.45 & 86.36 & 86.18 & 99.27 \\
Invalid inference & 99.75 & 98 & 69.25 & 60.75 & 41.5 & 59.25 \\
Predicate basics & 99.6 & 92.4 & 88.8 & 86 & 64 & 78.8 \\
Quantifier equivalence & 96.8 & 97.6 & 64.8 & 80 & 72.4 & 67.6 \\
Predicate inference & 100 & 100 & 100 & 86.29 & 82.86 & 98 \\
Multi-step reasoning & 97.5 & 93.5 & 70.5 & 97.5 & 88.5 & 95.5 \\
Relation logic & 99 & 93.5 & 97 & 97.5 & 70.35 & 88.5 \\
General (3,000) & 98.4 & 90.53 & 80.13 & 80.72 & 68.69 & 77.99 \\
Difficult (2,000) & 80.5 & 52.5 & 61.58 & 59.48 & 34 & 56.28 \\
\bottomrule
\end{tabular}
\caption{Logical-template-family-level accuracy for Qwen3-8B.}
\label{tab:app_qwen38b_summary}
\end{table*}

\begin{table*}[t]
\centering
\scriptsize
\setlength{\tabcolsep}{3.5pt}
\begin{tabular}{lrrrrrr}
\toprule
  & English & Ch-Std & Ch-Nat & Ch-Col & Ch-Rhet & Ch-Pert \\
\midrule
Basic connectives & 100 & 100 & 99.43 & 99.14 & 88.29 & 99.43 \\
Prop. equivalence & 94.67 & 60.22 & 60.22 & 34.22 & 48.67 & 42.89 \\
Prop. inference & 100 & 100 & 99.27 & 74.36 & 78.18 & 98.55 \\
Invalid inference & 100 & 99.25 & 88.5 & 84.5 & 82.5 & 88.5 \\
Predicate basics & 99.2 & 99.2 & 80.8 & 89.2 & 57.2 & 69.2 \\
Quantifier equivalence & 100 & 96.8 & 88 & 84 & 76.4 & 69.2 \\
Predicate inference & 100 & 100 & 100 & 85.43 & 85.43 & 90 \\
Multi-step reasoning & 100 & 100 & 52 & 68 & 87.5 & 87.5 \\
Relation logic & 100 & 100 & 94.97 & 71.86 & 90.45 & 85.93 \\
General (3,000) & 99.13 & 93.6 & 86.16 & 85.75 & 75.86 & 85.02 \\
Difficult (2,000) & 93.02 & 58.28 & 60.42 & 63.12 & 40.22 & 51.62 \\
\bottomrule
\end{tabular}
\caption{Logical-template-family-level accuracy for Qwen3-14B.}
\label{tab:app_qwen314b_summary}
\end{table*}

\begin{table*}[t]
\centering
\scriptsize
\setlength{\tabcolsep}{3.5pt}
\begin{tabular}{lrrrrrr}
\toprule
  & English & Ch-Std & Ch-Nat & Ch-Col & Ch-Rhet & Ch-Pert \\
\midrule
Basic connectives & 100 & 100 & 100 & 100 & 98.86 & 100 \\
Prop. equivalence & 99.33 & 78.67 & 78 & 90.89 & 94.44 & 85.78 \\
Prop. inference & 100 & 100 & 100 & 100 & 100 & 99.82 \\
Invalid inference & 93.75 & 92 & 84 & 79 & 79.75 & 95 \\
Predicate basics & 100 & 100 & 96.4 & 98.8 & 90 & 100 \\
Quantifier equivalence & 100 & 100 & 89.2 & 99.6 & 81.6 & 100 \\
Predicate inference & 100 & 100 & 100 & 100 & 100 & 100 \\
Multi-step reasoning & 100 & 100 & 99.5 & 100 & 95 & 97.5 \\
Relation logic & 100 & 100 & 100 & 100 & 98.5 & 100 \\
General (3,000) & 99.07 & 95.73 & 93.33 & 95.7 & 93.53 & 97 \\
Difficult (2,000) & 96.05 & 83.1 & 87.8 & 86.15 & 69.35 & 85.2 \\
\bottomrule
\end{tabular}
\caption{Logical-template-family-level accuracy for Qwen3-32B.}
\label{tab:app_qwen332b_summary}
\end{table*}

\begin{table*}[t]
\centering
\scriptsize
\setlength{\tabcolsep}{3.5pt}
\begin{tabular}{lrrrrrr}
\toprule
  & English & Ch-Std & Ch-Nat & Ch-Col & Ch-Rhet & Ch-Pert \\
\midrule
Basic connectives & 48.57 & 46.86 & 28.57 & 23.14 & 18 & 23.43 \\
Prop. equivalence & 19.11 & 0.8 & 0 & 0 & 0 & 1.11 \\
Prop. inference & 76.36 & 32 & 34 & 20.18 & 21.82 & 54.91 \\
Invalid inference & 91.5 & 100 & 100 & 100 & 100 & 100 \\
Predicate basics & 66.8 & 6 & 2.8 & 2.8 & 0 & 3.6 \\
Quantifier equivalence & 20 & 21.2 & 21.2 & 42.8 & 33.6 & 20 \\
Predicate inference & 59.14 & 50.57 & 53.71 & 31.14 & 31.71 & 34.57 \\
Multi-step reasoning & 94.5 & 69.5 & 63 & 55 & 36 & 25.5 \\
Relation logic & 61.5 & 63 & 82.5 & 48 & 52 & 47.5 \\
General (3,000) & 59.27 & 41.68 & 40.88 & 30.02 & 31.82 & 37.17 \\
Difficult (2,000) & 73.3 & 47.9 & 49.02 & 49.06 & 46.6 & 46.3 \\
\bottomrule
\end{tabular}
\caption{Logical-template-family-level accuracy for Ministral-3B.}
\label{tab:app_ministral3b_summary}
\end{table*}

\begin{table*}[t]
\centering
\scriptsize
\setlength{\tabcolsep}{3.5pt}
\begin{tabular}{lrrrrrr}
\toprule
  & English & Ch-Std & Ch-Nat & Ch-Col & Ch-Rhet & Ch-Pert \\
\midrule
Basic connectives & 99.71 & 100 & 98.96 & 99.43 & 74.57 & 98.29 \\
Prop. equivalence & 94.67 & 23.56 & 28.89 & 21.33 & 8 & 42.67 \\
Prop. inference & 99.45 & 87.27 & 90.91 & 83.09 & 77.82 & 97.27 \\
Invalid inference & 73.25 & 94.75 & 92.75 & 96 & 96.25 & 94.5 \\
Predicate basics & 98.8 & 84.8 & 85.6 & 78 & 80.4 & 84.4 \\
Quantifier equivalence & 80.4 & 48.8 & 76.4 & 76.8 & 70.4 & 67.6 \\
Predicate inference & 100 & 100 & 99.43 & 92.86 & 82.57 & 99.71 \\
Multi-step reasoning & 96 & 100 & 100 & 100 & 93.5 & 86.5 \\
Relation logic & 78 & 98 & 85 & 89.45 & 77 & 87.5 \\
General (3,000) & 92.03 & 79.83 & 82.35 & 79.23 & 70.57 & 84.2 \\
Difficult (2,000) & 68.55 & 64.55 & 77.05 & 72.4 & 49.25 & 72.55 \\
\bottomrule
\end{tabular}
\caption{Logical-template-family-level accuracy for Ministral-8B.}
\label{tab:app_ministral8b_summary}
\end{table*}

\begin{table*}[t]
\centering
\scriptsize
\setlength{\tabcolsep}{3.5pt}
\begin{tabular}{lrrrrrr}
\toprule
  & English & Ch-Std & Ch-Nat & Ch-Col & Ch-Rhet & Ch-Pert \\
\midrule
Basic connectives & 98.86 & 91.14 & 99.43 & 96.57 & 99.43 & 88.57 \\
Prop. equivalence & 98.22 & 78.22 & 34.44 & 81.56 & 43.33 & 35.56 \\
Prop. inference & 100 & 98.73 & 99.09 & 82.55 & 99.27 & 97.82 \\
Invalid inference & 98.5 & 99.5 & 97 & 85 & 90.5 & 99.75 \\
Predicate basics & 94 & 90 & 80 & 72 & 52.8 & 54.8 \\
Quantifier equivalence & 94.8 & 79.2 & 56.8 & 63.2 & 57.8 & 92.8 \\
Predicate inference & 100 & 100 & 100 & 98.29 & 99.43 & 99.71 \\
Multi-step reasoning & 99.5 & 97 & 83 & 93 & 57.5 & 93 \\
Relation logic & 98 & 99.5 & 98.5 & 90.5 & 74.5 & 92 \\
General (3,000) & 98.3 & 92.6 & 84.56 & 84.94 & 78.89 & 83.17 \\
Difficult (2,000) & 84.7 & 81 & 82.5 & 78.7 & 52.3 & 78.05 \\
\bottomrule
\end{tabular}
\caption{Logical-template-family-level accuracy for GLM-5.1.}
\label{tab:app_glm51_summary}
\end{table*}

\subsection{Fine-grained results grouped by logical-template family}
For each logical-template family, the fine-grained results are divided into two tables: one for the Qwen3 models and one for the Ministral and GLM models. Each cell lists English / Ch-Std / Ch-Nat / Ch-Col / Ch-Rhet / Ch-Pert accuracy (\%). This organization preserves all reported values while keeping each table readable.

\begin{table*}[t]
\centering
\scriptsize
\setlength{\tabcolsep}{3.2pt}
\renewcommand{\arraystretch}{1.12}

\begin{tabular}{lcccc}
\toprule
Subtype & Qwen3-0.6B & Qwen3-8B & Qwen3-14B & Qwen3-32B \\
\midrule
Atomic proposition & 100/100/100/100/100/100 & 100/100/100/100/100/100 & 100/100/100/100/100/100 & 100/100/100/100/100/100 \\
Negation & 0/100/100/100/100/100 & 100/100/100/100/100/100 & 100/100/100/100/100/96 & 100/100/100/100/92/100 \\
Conjunction & 100/100/100/100/100/100 & 100/100/100/100/100/0 & 100/100/100/100/100/100 & 100/100/100/100/100/100 \\
Disjunction & 100/100/100/100/100/100 & 100/100/100/100/100/82 & 100/100/100/98/24/100 & 100/100/100/100/100/100 \\
Exclusive disjunction & 100/100/100/100/100/100 & 100/4/94/100/74/100 & 100/100/100/100/98/100 & 100/100/100/100/100/100 \\
Implication & 100/100/100/100/100/100 & 100/100/100/100/96/100 & 100/100/100/96/96/100 & 100/100/100/100/100/100 \\
Biconditional & 100/100/100/100/100/100 & 100/100/94/100/98/100 & 100/100/96/100/100/100 & 100/100/100/100/100/100 \\
\bottomrule
\end{tabular}%

\caption{Fine-grained accuracy (\%) for Basic connectives. Results for the Qwen3 models. Each cell reports English / Ch-Std / Ch-Nat / Ch-Col / Ch-Rhet / Ch-Pert.}
\label{tab:app_fg_basic_qwen}
\end{table*}

\begin{table*}[t]
\centering
\scriptsize
\setlength{\tabcolsep}{3.2pt}
\renewcommand{\arraystretch}{1.12}

\begin{tabular}{lccc}
\toprule
Subtype & Ministral-3B & Ministral-8B & GLM-5.1 \\
\midrule
Atomic proposition & 40/0/0/0/0/0 & 98/100/100/100/100/96 & 100/100/100/100/100/100 \\
Negation & 28/100/100/100/100/100 & 100/100/100/100/100/100 & 100/100/100/98/100/100 \\
Conjunction & 48/46/0/0/8/0 & 100/100/100/100/100/92 & 100/100/98/100/100/20 \\
Disjunction & 12/10/2/0/0/2 & 100/100/100/100/72/100 & 100/100/100/80/100/100 \\
Exclusive disjunction & 36/0/0/30/18/0 & 100/100/100/100/100/100 & 92/38/100/100/96/100 \\
Implication & 100/96/80/32/0/44 & 100/100/92/96/36/100 & 100/100/98/100/100/100 \\
Biconditional & 76/76/18/0/0/16 & 100/100/100/100/14/100 & 100/100/100/98/100/100 \\
\bottomrule
\end{tabular}%

\caption{Fine-grained accuracy (\%) for Basic connectives. Results for the Ministral and GLM models. Each cell reports English / Ch-Std / Ch-Nat / Ch-Col / Ch-Rhet / Ch-Pert.}
\label{tab:app_fg_basic_other}
\end{table*}

\begin{table*}[t]
\centering
\scriptsize
\setlength{\tabcolsep}{3.2pt}
\renewcommand{\arraystretch}{1.12}

\begin{tabular}{lcccc}
\toprule
Subtype & Qwen3-0.6B & Qwen3-8B & Qwen3-14B & Qwen3-32B \\
\midrule
Idempotence & 100/100/100/100/100/100 & 100/76/48/68/34/74 & 100/50/60/42/44/36 & 100/72/90/86/100/94 \\
Commutativity & 100/100/100/100/100/100 & 100/96/58/88/76/88 & 100/98/90/74/86/74 & 100/100/84/100/100/100 \\
Associativity & 100/100/100/100/100/100 & 90/0/38/0/0/38 & 94/18/28/0/48/18 & 100/50/56/52/94/62 \\
Distributivity & 100/100/100/100/100/100 & 72/2/8/0/0/24 & 98/18/30/2/28/24 & 100/88/66/96/94/68 \\
De Morgan & 100/100/100/100/100/100 & 84/82/32/34/28/60 & 66/72/70/50/40/68 & 94/82/76/100/90/80 \\
Double negation & 100/100/100/100/100/100 & 100/100/50/100/40/72 & 100/88/72/58/42/32 & 100/80/90/100/100/100 \\
Conditional law & 100/100/100/100/100/100 & 100/100/60/96/58/60 & 100/98/84/42/60/46 & 100/64/74/96/100/88 \\
Contraposition & 100/100/100/100/100/100 & 100/68/40/68/30/60 & 100/52/56/38/48/38 & 100/100/76/--/100/94 \\
Biconditional law & 100/100/100/100/100/100 & 96/32/36/34/10/50 & 94/48/52/2/42/50 & 100/72/90/88/90/86 \\
\bottomrule
\end{tabular}%

\caption{Fine-grained accuracy (\%) for Propositional equivalence. Results for the Qwen3 models. Each cell reports English / Ch-Std / Ch-Nat / Ch-Col / Ch-Rhet / Ch-Pert.}
\label{tab:app_fg_propeq_qwen}
\end{table*}

\begin{table*}[t]
\centering
\scriptsize
\setlength{\tabcolsep}{3.2pt}
\renewcommand{\arraystretch}{1.12}

\begin{tabular}{lccc}
\toprule
Subtype & Ministral-3B & Ministral-8B & GLM-5.1 \\
\midrule
Idempotence & 16/0/0/0/0/0 & 100/30/20/20/4/34 & 94/46/28/50/26/24 \\
Commutativity & 62/8/0/0/0/10 & 100/76/70/86/40/82 & 100/90/54/88/78/54 \\
Associativity & 0/0/0/0/0/0 & 86/0/20/0/2/22 & 100/54/8/86/40/18 \\
Distributivity & 18/0/0/0/0/0 & 96/2/6/0/6/32 & 100/72/12/52/34/6 \\
De Morgan & 0/0/0/0/0/0 & 74/16/18/2/0/38 & 94/88/46/100/52/58 \\
Double negation & 0/0/0/0/0/0 & 96/8/8/0/0/34 & 100/100/42/100/50/30 \\
Conditional law & 12/0/0/0/0/0 & 100/44/20/52/10/60 & 96/66/18/74/30/38 \\
Contraposition & 8/0/0/0/0/0 & 100/14/26/12/2/52 & 100/98/52/96/48/52 \\
Biconditional law & 56/0/0/0/0/0 & 100/22/18/20/8/30 & 100/90/50/88/32/40 \\
\bottomrule
\end{tabular}

\caption{Fine-grained accuracy (\%) for Propositional equivalence. Results for the Ministral and GLM models. Each cell reports English / Ch-Std / Ch-Nat / Ch-Col / Ch-Rhet / Ch-Pert.}
\label{tab:app_fg_propeq_other}
\end{table*}

\begin{table*}[t]
\centering
\scriptsize
\setlength{\tabcolsep}{3.2pt}
\renewcommand{\arraystretch}{1.12}

\begin{tabular}{lcccc}
\toprule
Subtype & Qwen3-0.6B & Qwen3-8B & Qwen3-14B & Qwen3-32B \\
\midrule
Modus ponens & 100/100/100/100/100/100 & 100/100/100/100/100/100 & 100/100/100/100/100/100 & 100/100/100/100/100/100 \\
Modus tollens & 100/100/100/100/100/100 & 100/88/52/98/100/98 & 100/100/92/20/4/96 & 100/100/100/100/100/100 \\
Hypothetical syllogism & 100/100/100/100/100/100 & 100/100/100/100/100/100 & 100/100/100/100/92/100 & 100/100/100/100/100/100 \\
Disjunctive syllogism & 100/100/100/100/100/100 & 100/100/100/100/100/100 & 100/100/100/100/100/100 & 100/100/100/100/100/100 \\
Addition & 100/100/100/100/100/100 & 100/100/100/100/100/100 & 100/100/100/100/100/100 & 100/100/100/100/100/100 \\
Simplification & 100/100/100/100/100/100 & 100/96/98/92/100/100 & 100/100/100/52/100/92 & 100/100/100/100/100/98 \\
Conjunction introduction & 100/100/100/100/100/100 & 100/98/100/100/100/100 & 100/100/100/100/100/100 & 100/100/100/100/100/100 \\
Resolution & 100/100/100/100/100/100 & 100/100/100/40/48/94 & 100/100/100/30/0/96 & 100/100/100/100/100/100 \\
Disjunction elimination & 100/100/100/100/100/100 & 100/100/100/100/100/100 & 100/100/100/100/100/100 & 100/100/100/100/100/100 \\
Biconditional introduction & 100/100/100/100/100/100 & 100/100/100/20/0/100 & 100/100/100/16/64/100 & 100/100/100/100/100/100 \\
Biconditional elimination & 100/100/100/100/100/100 & 100/100/100/100/100/100 & 100/100/100/100/100/100 & 100/100/100/100/100/100 \\
\bottomrule
\end{tabular}%

\caption{Fine-grained accuracy (\%) for Propositional inference. Results for the Qwen3 models. Each cell reports English / Ch-Std / Ch-Nat / Ch-Col / Ch-Rhet / Ch-Pert.}
\label{tab:app_fg_propinf_qwen}
\end{table*}

\begin{table*}[t]
\centering

\fontsize{8.5pt}{9pt}\selectfont

\setlength{\tabcolsep}{3.2pt}
\renewcommand{\arraystretch}{1.12}

\begin{tabular}{lccc}
\toprule
Subtype & Ministral-3B & Ministral-8B & GLM-5.1 \\
\midrule
Modus ponens & 100/60/96/18/50/100 & 100/100/100/94/100/100 & 100/100/100/100/100/100 \\
Modus tollens & 4/0/0/0/0/0 & 100/48/2/68/8/72 & 100/100/96/100/100/78 \\
Hypothetical syllogism & 100/80/94/62/54/100 & 100/100/100/94/96/100 & 100/100/100/98/100/100 \\
Disjunctive syllogism & 98/2/0/0/58/26 & 100/100/100/98/100/100 & 100/100/100/100/100/100 \\
Addition & 76/6/44/0/0/70 & 100/100/100/100/100/100 & 100/100/100/100/100/100 \\
Simplification & 30/0/4/0/2/0 & 96/64/98/80/100/100 & 100/86/94/90/100/98 \\
Conjunction introduction & 56/14/44/74/0/62 & 98/78/100/100/100/100 & 100/100/100/100/100/100 \\
Resolution & 76/4/0/0/0/4 & 100/100/100/68/26/100 & 100/100/100/12/98/98 \\
Disjunction elimination & 100/74/22/24/10/74 & 100/100/100/100/100/100 & 100/100/100/100/100/100 \\
Biconditional introduction & 100/46/32/0/4/82 & 100/92/100/12/26/98 & 100/100/100/8/94/100 \\
Biconditional elimination & 100/66/34/44/62/86 & 100/100/98/100/100/100 & 100/100/100/100/100/100 \\
\bottomrule
\end{tabular}

\caption{Fine-grained accuracy (\%) for Propositional inference. Results for the Ministral and GLM models. Each cell reports English / Ch-Std / Ch-Nat / Ch-Col / Ch-Rhet / Ch-Pert.}
\label{tab:app_fg_propinf_other}
\end{table*}

\begin{table*}[t]
\centering
\fontsize{8.5pt}{9pt}\selectfont

\setlength{\tabcolsep}{3.2pt}
\renewcommand{\arraystretch}{1.12}

\begin{tabular}{lcccc}
\toprule
Subtype & Qwen3-0.6B & Qwen3-8B & Qwen3-14B & Qwen3-32B \\
\midrule
Affirming consequent & 0/0/0/0/0/0 & 98/98/62/46/44/54 & 100/96/80/78/96/100 & 100/96/90/96/96/96 \\
Denying antecedent & 0/0/0/0/0/0 & 100/86/56/84/54/44 & 100/98/78/96/76/88 & 96/84/70/42/82/100 \\
Affirming disjunct & 0/0/0/0/0/0 & 100/100/94/30/22/60 & 100/100/100/74/60/84 & 54/100/90/90/92/96 \\
Illicit conversion & 0/0/0/0/0/0 & 100/100/98/40/18/58 & 100/100/100/62/62/94 & 100/58/82/76/78/98 \\
Existential fallacy & 0/0/0/0/0/0 & 100/100/74/100/100/74 & 100/100/86/100/100/96 & 100/100/40/48/26/82 \\
Illicit major & 0/0/0/0/0/0 & 100/100/0/50/4/4 & 10/100/86/100/68/52 & 100/98/100/100/100/88 \\
Illicit minor & 0/0/0/0/0/0 & 100/100/90/84/40/98 & 100/100/100/96/100/100 & 100/100/100/100/100/100 \\
Undistributed middle & 0/0/0/0/0/0 & 100/100/80/52/50/82 & 100/100/78/70/98/94 & 100/100/100/80/64/100 \\
\bottomrule
\end{tabular}%

\caption{Fine-grained accuracy (\%) for Invalid inference / fallacy. Results for the Qwen3 models. Each cell reports English / Ch-Std / Ch-Nat / Ch-Col / Ch-Rhet / Ch-Pert.}
\label{tab:app_fg_invalid_qwen}
\end{table*}

\begin{table*}[t]
\centering
\fontsize{7.5pt}{9pt}\selectfont

\setlength{\tabcolsep}{3.2pt}
\renewcommand{\arraystretch}{1.12}

\begin{tabular}{lccc}
\toprule
Subtype & Ministral-3B & Ministral-8B & GLM-5.1 \\
\midrule
Affirming consequent & 72/100/100/100/100/100 & 30/84/96/94/84/98 & 100/100/100/62/88/100 \\
Denying antecedent & 100/100/100/100/100/100 & 68/82/98/88/100/100 & 100/96/96/100/98/100 \\
Affirming disjunct & 92/100/100/100/100/100 & 78/94/92/100/100/100 & 100/100/98/48/82/100 \\
Illicit conversion & 100/100/100/100/100/100 & 100/100/100/100/98/100 & 100/100/100/100/78/100 \\
Existential fallacy & 72/100/100/100/100/100 & 24/100/58/88/98/94 & 88/100/98/100/86/100 \\
Illicit major & 98/100/100/100/100/100 & 88/100/98/100/100/66 & 100/100/84/100/96/98 \\
Illicit minor & 100/100/100/100/100/100 & 100/100/100/100/100/100 & 100/100/100/100/96/100 \\
Undistributed middle & 98/100/100/100/100/100 & 98/98/100/98/90/98 & 100/100/100/70/100/100 \\
\bottomrule
\end{tabular}%

\caption{Fine-grained accuracy (\%) for Invalid inference / fallacy. Results for the Ministral and GLM models. Each cell reports English / Ch-Std / Ch-Nat / Ch-Col / Ch-Rhet / Ch-Pert.}
\label{tab:app_fg_invalid_other}
\end{table*}

\begin{table*}[t]
\centering
\scriptsize
\setlength{\tabcolsep}{3.2pt}
\renewcommand{\arraystretch}{1.12}

\begin{tabular}{lcccc}
\toprule
Subtype & Qwen3-0.6B & Qwen3-8B & Qwen3-14B & Qwen3-32B \\
\midrule
Unary predicate & 100/100/100/100/100/100 & 100/100/100/60/98/100 & 100/100/100/98/64/100 & 100/100/100/98/100/100 \\
Binary predicate & 100/100/100/100/100/100 & 100/100/72/86/28/88 & 100/100/16/88/100/100 & 100/100/100/96/98/100 \\
Universal quantifier & 100/100/100/100/100/100 & 100/100/76/88/78/32 & 10/100/90/72/6/18 & 100/100/82/100/52/100 \\
Existential quantifier & 100/100/100/100/100/100 & 100/64/98/100/28/92 & 100/98/100/94/100/98 & 100/100/100/100/100/100 \\
Unique existence & 100/100/100/100/100/100 & 98/98/98/98/88/82 & 96/98/98/94/16/30 & 100/100/100/100/100/100 \\
\bottomrule
\end{tabular}%

\caption{Fine-grained accuracy (\%) for Predicate basics. Results for the Qwen3 models. Each cell reports English / Ch-Std / Ch-Nat / Ch-Col / Ch-Rhet / Ch-Pert.}
\label{tab:app_fg_predbasic_qwen}
\end{table*}

\begin{table*}[t]
\centering
\scriptsize
\setlength{\tabcolsep}{3.2pt}
\renewcommand{\arraystretch}{1.12}
\begin{tabular}{lccc}
\toprule
Subtype & Ministral-3B & Ministral-8B & GLM-5.1 \\
\midrule
Unary predicate & 94/6/2/0/0/8 & 100/100/100/84/100/98 & 100/98/100/94/94/52 \\
Binary predicate & 78/20/2/4/0/0 & 100/100/66/86/90/100 & 100/100/38/78/44/28 \\
Universal quantifier & 78/4/0/0/0/0 & 100/98/64/54/26/56 & 100/96/64/82/56/78 \\
Existential quantifier & 76/0/10/8/0/6 & 100/52/100/76/100/100 & 100/84/100/98/54/100 \\
Unique existence & 8/0/0/2/0/4 & 94/74/98/90/86/68 & 70/72/98/8/16/16 \\
\bottomrule
\end{tabular}%

\caption{Fine-grained accuracy (\%) for Predicate basics. Results for the Ministral and GLM models. Each cell reports English / Ch-Std / Ch-Nat / Ch-Col / Ch-Rhet / Ch-Pert.}
\label{tab:app_fg_predbasic_other}
\end{table*}

\begin{table*}[t]
\centering
\scriptsize
\setlength{\tabcolsep}{3.2pt}
\renewcommand{\arraystretch}{1.12}

\begin{tabular}{lcccc}
\toprule
Subtype & Qwen3-0.6B & Qwen3-8B & Qwen3-14B & Qwen3-32B \\
\midrule
Quantifier negation & 100/100/100/100/100/100 & 84/96/88/82/98/98 & 100/100/98/82/92/100 & 100/100/90/98/100/100 \\
Universal negation & 100/100/100/100/100/100 & 100/100/60/100/100/38 & 100/100/96/100/100/24 & 100/100/72/100/100/100 \\
Existential disjunction & 100/100/100/100/100/100 & 100/100/78/100/98/90 & 100/88/96/100/92/100 & 100/100/94/100/100/100 \\
Invalid quantifier distribution & 0/0/0/0/0/0 & 100/98/52/44/66/60 & 100/100/52/62/98/72 & 100/100/100/100/100/100 \\
Quantifier movement & 100/100/100/100/100/100 & 100/94/46/74/0/52 & 100/96/98/76/0/50 & 100/100/90/100/8/100 \\
\bottomrule
\end{tabular}%

\caption{Fine-grained accuracy (\%) for Quantifier equivalence. Results for the Qwen3 models. Each cell reports English / Ch-Std / Ch-Nat / Ch-Col / Ch-Rhet / Ch-Pert.}
\label{tab:app_fg_quanteq_qwen}
\end{table*}

\begin{table*}[t]
\centering
\scriptsize
\setlength{\tabcolsep}{3.2pt}
\renewcommand{\arraystretch}{1.12}

\begin{tabular}{lccc}
\toprule
Subtype & Ministral-3B & Ministral-8B & GLM-5.1 \\
\midrule
Quantifier negation & 0/0/0/0/0/0 & 100/60/70/50/52/98 & 98/98/80/70/78/92 \\
Universal negation & 2/0/0/28/68/0 & 100/66/54/96/100/28 & 98/100/66/100/100/94 \\
Existential disjunction & 0/6/0/86/0/0 & 100/8/90/100/100/84 & 100/80/28/58/2/98 \\
Invalid quantifier distribution & 78/100/100/100/100/100 & 4/100/100/86/100/96 & 100/100/58/78/70/100 \\
Quantifier movement & 20/0/6/0/0/0 & 98/10/68/54/0/32 & 78/18/52/10/10/90 \\
\bottomrule
\end{tabular}%

\caption{Fine-grained accuracy (\%) for Quantifier equivalence. Results for the Ministral and GLM models. Each cell reports English / Ch-Std / Ch-Nat / Ch-Col / Ch-Rhet / Ch-Pert.}
\label{tab:app_fg_quanteq_other}
\end{table*}

\begin{table*}[t]
\centering
\scriptsize
\setlength{\tabcolsep}{3.2pt}
\renewcommand{\arraystretch}{1.12}

\begin{tabular}{lcccc}
\toprule
Subtype & Qwen3-0.6B & Qwen3-8B & Qwen3-14B & Qwen3-32B \\
\midrule
Universal instantiation & 100/100/100/100/100/100 & 100/100/100/100/100/100 & 100/100/100/90/100/98 & 100/100/100/100/100/100 \\
Existential generalization & 100/100/100/100/100/100 & 100/100/100/100/100/100 & 100/100/100/100/100/100 & 100/100/100/100/100/100 \\
Universal generalization & 0/0/0/0/0/0 & 100/100/100/6/80/100 & 100/100/100/100/100/100 & 100/100/100/100/100/100 \\
Predicate MP & 100/100/100/100/100/100 & 100/100/100/100/100/100 & 100/100/100/100/100/98 & 100/100/100/100/100/100 \\
Predicate MT & 100/100/100/100/100/100 & 100/100/100/98/0/86 & 100/100/100/8/0/34 & 100/100/100/100/100/100 \\
Predicate transitive inference & 100/100/100/100/100/100 & 100/100/100/100/100/100 & 100/100/100/100/98/100 & 100/100/100/100/100/100 \\
Existential instance inference & 100/100/100/100/100/100 & 100/100/100/100/100/100 & 100/100/100/100/100/100 & 100/100/100/100/100/100 \\
\bottomrule
\end{tabular}%

\caption{Fine-grained accuracy (\%) for Predicate inference. Results for the Qwen3 models. Each cell reports English / Ch-Std / Ch-Nat / Ch-Col / Ch-Rhet / Ch-Pert.}
\label{tab:app_fg_predinf_qwen}
\end{table*}

\begin{table*}[t]
\centering
\scriptsize
\setlength{\tabcolsep}{3.2pt}
\renewcommand{\arraystretch}{1.12}

\begin{tabular}{lccc}
\toprule
Subtype & Ministral-3B & Ministral-8B & GLM-5.1 \\
\midrule
Universal instantiation & 100/90/64/12/0/22 & 100/100/100/100/96/100 & 100/100/100/100/100/100 \\
Existential generalization & 0/0/0/0/28/0 & 100/100/100/100/100/100 & 100/100/100/100/100/100 \\
Universal generalization & 98/100/100/100/100/100 & 100/100/100/100/100/100 & 100/100/100/90/98/98 \\
Predicate MP & 100/90/46/14/2/42 & 100/100/100/96/82/98 & 100/100/100/98/100/100 \\
Predicate MT & 0/0/0/0/0/0 & 100/100/96/54/0/100 & 100/100/100/100/98/100 \\
Predicate transitive inference & 88/62/76/--/92/54 & 100/100/100/100/100/100 & 100/100/100/100/100/100 \\
Existential instance inference & 28/12/90/24/0/24 & 100/100/100/100/100/100 & 100/100/100/100/100/100 \\
\bottomrule
\end{tabular}%

\caption{Fine-grained accuracy (\%) for Predicate inference. Results for the Ministral and GLM models. Each cell reports English / Ch-Std / Ch-Nat / Ch-Col / Ch-Rhet / Ch-Pert.}
\label{tab:app_fg_predinf_other}
\end{table*}

\begin{table*}[t]
\centering
\scriptsize
\setlength{\tabcolsep}{3.2pt}
\renewcommand{\arraystretch}{1.12}

\begin{tabular}{lcccc}
\toprule
Subtype & Qwen3-0.6B & Qwen3-8B & Qwen3-14B & Qwen3-32B \\
\midrule
Linear chain & 100/100/100/100/100/100 & 100/100/52/98/60/92 & 100/100/18/32/50/52 & 100/100/98/100/80/90 \\
Conjunctive condition chain & 100/100/100/100/100/100 & 100/100/100/94/100/100 & 100/100/100/100/100/100 & 100/100/100/100/100/100 \\
Branch reasoning & 0/0/0/0/0/0 & 90/74/30/98/100/90 & 100/100/0/8/100/98 & 100/100/100/100/100/100 \\
Multi-premise reasoning & 100/100/100/100/100/100 & 100/100/100/100/94/100 & 100/100/90/100/100/100 & 100/100/100/100/100/100 \\
\bottomrule
\end{tabular}%

\caption{Fine-grained accuracy (\%) for Multi-step reasoning. Results for the Qwen3 models. Each cell reports English / Ch-Std / Ch-Nat / Ch-Col / Ch-Rhet / Ch-Pert.}
\label{tab:app_fg_multistep_qwen}
\end{table*}

\begin{table*}[t]
\centering
\scriptsize
\setlength{\tabcolsep}{3.2pt}
\renewcommand{\arraystretch}{1.12}

\begin{tabular}{lccc}
\toprule
Subtype & Ministral-3B & Ministral-8B & GLM-5.1 \\
\midrule
Linear chain & 100/100/36/22/6/2 & 100/100/100/100/82/56 & 100/100/32/76/28/92 \\
Conjunctive condition chain & 100/58/100/36/36/0 & 100/100/100/100/100/100 & 100/100/100/96/90/98 \\
Branch reasoning & 96/100/100/100/100/100 & 84/100/100/100/100/100 & 98/88/100/100/100/98 \\
Multi-premise reasoning & 82/20/16/62/2/0 & 100/100/100/100/92/90 & 100/100/100/100/12/82 \\
\bottomrule
\end{tabular}%

\caption{Fine-grained accuracy (\%) for Multi-step reasoning. Results for the Ministral and GLM models. Each cell reports English / Ch-Std / Ch-Nat / Ch-Col / Ch-Rhet / Ch-Pert.}
\label{tab:app_fg_multistep_other}
\end{table*}

\begin{table*}[t]
\centering
\scriptsize
\setlength{\tabcolsep}{3.2pt}
\renewcommand{\arraystretch}{1.12}

\begin{tabular}{lcccc}
\toprule
Subtype & Qwen3-0.6B & Qwen3-8B & Qwen3-14B & Qwen3-32B \\
\midrule
Transitive relation & 100/100/100/100/100/100 & 100/100/100/100/98/100 & 100/100/100/84/90/100 & 100/100/100/100/100/100 \\
Non-transitive relation & 0/0/0/0/0/0 & 96/88/88/90/83.67/80 & 100/100/80/100/100/79.59 & 100/100/100/100/100/100 \\
Symmetric relation & 100/100/100/100/100/100 & 100/100/100/100/0/100 & 100/100/100/100/90/100 & 100/100/100/100/94/100 \\
Antisymmetric relation & 100/100/100/100/100/100 & 100/86/100/100/100/74 & 100/100/100/4/82/64 & 100/100/100/100/100/100 \\
\bottomrule
\end{tabular}%

\caption{Fine-grained accuracy (\%) for Relation logic. Results for the Qwen3 models. Each cell reports English / Ch-Std / Ch-Nat / Ch-Col / Ch-Rhet / Ch-Pert.}
\label{tab:app_fg_relation_qwen}
\end{table*}

\begin{table*}[t]
\centering
\scriptsize
\setlength{\tabcolsep}{3.2pt}
\renewcommand{\arraystretch}{1.12}

\begin{tabular}{lccc}
\toprule
Subtype & Ministral-3B & Ministral-8B & GLM-5.1 \\
\midrule
Transitive relation & 100/94/98/84/84/88 & 100/100/100/98/100/100 & 100/100/100/92/100/100 \\
Non-transitive relation & 42/100/100/100/100/100 & 12/92/40/100/76/74 & 92/98/94/92/94/92 \\
Symmetric relation & 62/14/74/10/0/0 & 100/100/100/100/34/100 & 100/100/100/100/4/100 \\
Antisymmetric relation & 42/46/60/0/26/4 & 100/100/100/60/100/76 & 100/100/100/78/100/76 \\
\bottomrule
\end{tabular}%

\caption{Fine-grained accuracy (\%) for Relation logic. Results for the Ministral and GLM models. Each cell reports English / Ch-Std / Ch-Nat / Ch-Col / Ch-Rhet / Ch-Pert.}
\label{tab:app_fg_relation_other}
\end{table*}

\subsection{Enhancement experiments}
\begin{table*}[t]
\centering
\scriptsize
\setlength{\tabcolsep}{3.5pt}

\begin{tabular}{llrrrrr}
\toprule
Model & Condition & Invalid quantifier dist. & Existential fallacy & Illicit major & Illicit minor & Branch reasoning \\
\midrule
Qwen3-8B & Original English & 62 & 94 & 90 & 92 & 10 \\
 & Enhanced English & 100 & 100 & 100 & 100 & 100 \\
 & Original translation & 52 & 98 & 62 & 98 & 52 \\
 & Enhanced translation & 98 & 100 & 100 & 100 & 100 \\
\midrule
Qwen3-14B & Original English & 100 & 100 & 96 & 98 & 0 \\
 & Enhanced English & 100 & 100 & 100 & 100 & 100 \\
 & Original translation & 52 & 100 & 100 & 100 & 40 \\
 & Enhanced translation & 100 & 100 & 100 & 100 & 100 \\
\bottomrule
\end{tabular}%

\caption{Enhancement experiments for Qwen3-8B and Qwen3-14B.}
\label{tab:enh_qwen_combined}
\end{table*}

\end{document}